\documentclass[letterpaper, 10 pt, conference]{ieeeconf}
\IEEEoverridecommandlockouts
\usepackage{url}
\usepackage{wrapfig}
\usepackage{graphicx}
\usepackage{amsmath, amssymb}
\usepackage{subcaption}
\usepackage{booktabs}
\usepackage{algorithm}
\usepackage{algpseudocode}
\usepackage{xcolor}

\title{\LARGE \bf Data-Efficient Multitask DAgger}
\author{Haotian Fu$^{1*}$, Ran Gong$^2$, Xiaohan Zhang$^2$, Maria Vittoria Minniti$^2$, Jigarkumar Patel$^2$, Karl Schmeckpeper$^2$ \\
$^1$Brown University 
$^2$Robotics and AI Institute
\thanks{$^*$Work done as an intern at Robotics and AI Institute}
}
\date{}

\begin{document}
\maketitle
\thispagestyle{empty}
\pagestyle{empty}

\begin{abstract}
Generalist robot policies that can perform many tasks typically require extensive expert data or simulations for training. In this work, we propose a novel Data-Efficient \emph{multitask DAgger} framework that distills a single multitask policy from multiple task-specific expert policies. Our approach significantly increases the overall task success rate by actively focusing on tasks where the multitask policy underperforms. The core of our method is a performance-aware scheduling strategy that tracks how much each task's learning process benefits from the amount of data, using a Kalman filter-based estimator to robustly decide how to allocate additional demonstrations across tasks. We validate our approach on MetaWorld, as well as a suite of diverse drawer-opening tasks in IsaacLab. The resulting policy attains high performance across all tasks while using substantially fewer expert demonstrations, and the visual policy learned with our method in simulation shows better performance than naive DAgger and Behavior Cloning when transferring zero-shot to a real robot without using real data.
\end{abstract}

\section{Introduction}
Recent progress in robot learning has produced multitask policies \cite{black2410pi0, intelligence2025pi_, huang2024embodied, pmlr-v270-kim25c, mees2023grounding} capable of performing many manipulation tasks, moving towards the goal of foundation models for robotics. A major challenge in training such multitask policies is the requirement of large and diverse demonstration datasets covering all tasks of interest. Prior works have collected massive human teleoperated demonstration sets or autonomously generated data for dozens of tasks, but at the cost of considerable time and effort. Making multitask imitation learning more data-efficient is crucial for scaling up the number of tasks a single policy can learn.

In imitation learning, Dataset Aggregation (DAgger)~\cite{ross2011dagger} is a powerful technique to improve policy performance by iteratively collecting new expert data on states where a learned policy is likely to make mistakes. However, applying DAgger across many tasks raises a key question: how should we distribute the limited budget of data collection queries among the tasks? A naive strategy might allocate data uniformly or equally to each task. This simplistic approach is inefficient because it can waste expert effort on tasks where the policy already performs well, while not providing enough data on harder tasks. Conversely, an ideal strategy would focus on tasks that are currently challenging for the policy and where additional data would yield the greatest improvement.

\begin{figure*}[t]
    \centering
    \includegraphics[width=0.92\linewidth]{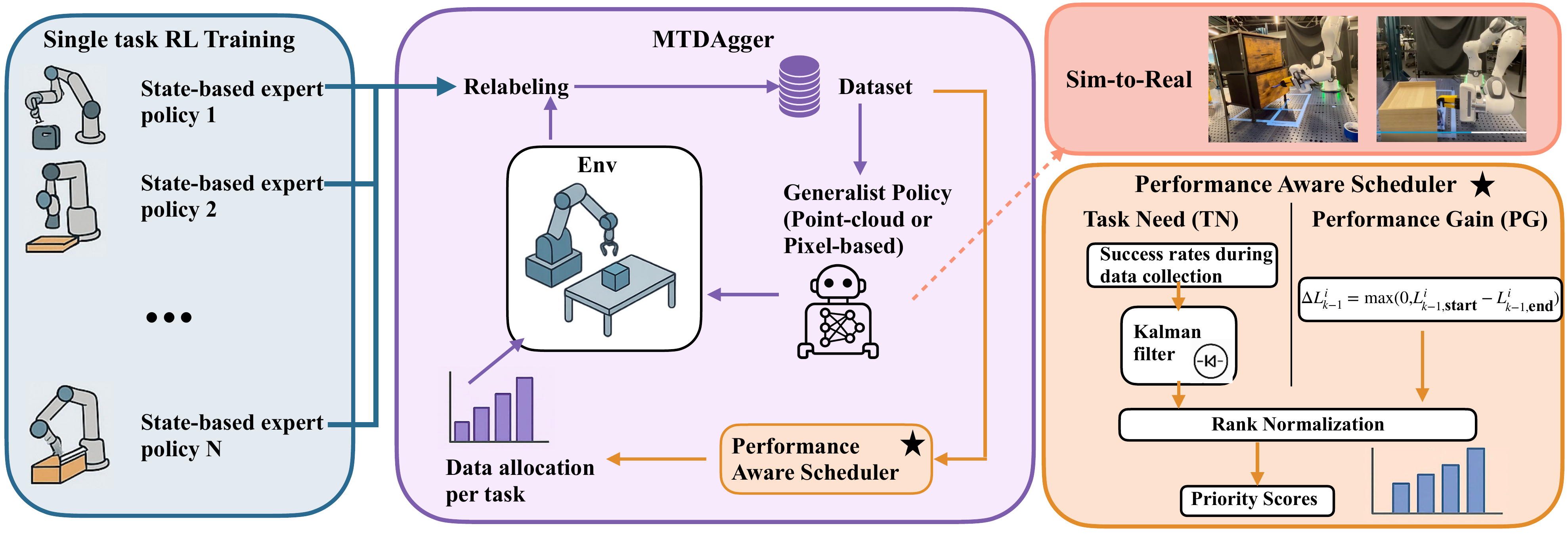}
    \caption{\textbf{Overview of our data-efficient multitask DAgger framework.} Multiple task-specific expert policies (blue) provide demonstrations to train a single multitask policy. In each DAgger iteration, the multitask policy is deployed on all tasks and a \emph{performance-aware scheduler} (orange) uses the Task Need (success rate) and Performance Gain (loss improvement) metrics to prioritize which tasks should receive new expert demonstrations. This focused data collection yields a high-performing generalist policy with far fewer demonstrations. The multitask policy trained in simulation can be directly transferred to real-world tasks.}
    \label{fig:overview}
\end{figure*}

In this paper, we introduce a \emph{data-efficient multitask DAgger} framework aiming to learn a multitask policy with much less data collection cost. We show the overview of the method in Figure~\ref{fig:overview}. We first train multiple task-specific expert policies (one per task) via reinforcement learning on state-based observations. Then our approach distills these experts into a single vision-based multitask policy (using point cloud or RGB inputs) via iterative imitation learning. Crucially, instead of querying each expert equally, our method monitors the multitask policy’s performance on each task and strategically biases data collection towards tasks that need it most. We design a performance-aware scheduler that uses one of two alternative metrics to prioritize tasks: \textbf{Task Need (TN)} - defined as the current success rate of the multitask policy on that task measured online during data collection (No additional evaluation phase is required to obtain TN), or \textbf{Performance Gain (PG)} - defined as the reduction in the task's behavior cloning loss after the latest training update, which reflects how much the policy's performance on that task improves with new data.  We normalize and combine these metrics into a priority score for each task. For the TN metric, we update the success-rate estimates using a Kalman filter to smooth out noise, enabling the scheduler to more reliably select which tasks get additional demonstrations in each DAgger iteration.

To summarize, our key contributions are as follows:
\begin{itemize}
    \item We propose a novel \textbf{data-efficient multitask DAgger framework} that distills a single multitask policy from multiple experts, significantly reducing the amount of expert-provided data needed per task.
    \item We introduce a \textbf{performance-aware data collection strategy} that tracks success rates and learning progress for each task, using a Kalman filter-based estimator to decide how to allocate new expert demonstrations across tasks for maximal performance gain.
    \item We present \textbf{simulation results} on MetaWorld and IsaacSim. The multitask policy outperforms behavior cloning baselines, and our performance-aware data scheduling helps the policy reach higher success rates with less data compared to a uniform DAgger strategy.
    \item We demonstrate \textbf{sim-to-real transfer}: the learned visual multitask policy, trained with our proposed method in simulation, exhibits better real world performance, especially on unseen objects. Robot demos can be found at \url{https://sites.google.com/brown.edu/mtdagger/home}.
\end{itemize}

\section{Related Work}

\textbf{MultiTask Imitation Learning.} 
Learning generalist policies has been explored in a number of recent works~\cite{singh2020scalable, mandlekar2020learning,jang2022bc-z,brohan2022rt1,zitkovich2023rt2,driess2023palme,haldar2024baku, black2410pi0, DBLP:conf/icml/FuSS0RCY24, intelligence2025pi_, team2025gemini, wan2024lotus}. 
While some of these models were pretrained on image or vision-question answering datasets~\cite{driess2023palme,zitkovich2023rt2},
all of them require some robot-specific data.  This data frequently comes from large-scale human teleoperation in the real world~\cite{ebert2021bridge,jang2022bc-z,o2024open}.  
On the other hand, our work follows the footsteps of other multitask policies trained with simulated data~\cite{jiang2022vima,bjorck2025gr00t}. However, rather than uniformly collecting a large number of trajectories, it intelligently samples from a number of expert policies to best improve the policy performance. Training generalist agents that can perform many tasks is a central goal in robotics. Our work also connects to curriculum learning~\cite{DBLP:conf/icml/BengioLCW09, DBLP:conf/icml/AndreasKL17, DBLP:conf/atal/IcarteKVM18}.

\textbf{Interactive Imitation Learning Algorithms.} 
Our approach builds on the DAgger algorithm~\cite{ross2011dagger}, which established the benefit of iterative expert feedback to correct compounding errors in learned policies. 
A key limitation of DAgger is the high cost of querying the expert at every timestep. Several works~\cite{DBLP:conf/iros/MendaDK19, DBLP:conf/rss/ZhangCK024} propose to query the expert when the policy is most uncertain. For example, \cite{zhang2016query, DBLP:journals/corr/abs-2502-13519} focus on preventing mistakes.
  Other works have learned failure predictors~\cite{zhang2016query}, used disagreement across an ensemble of learners~\cite{haridas2021dadagger}, or leveraged model-uncertainty~\cite{kelly2019hgdagger} to choose when to query the expert.
Like these works, we seek to limit expert intervention; however, instead of a binary query decision for a single task, our framework must decide \emph{which} task’s expert to query at each step. To our knowledge, adaptively allocating a demonstration budget across \textbf{multiple tasks} within a DAgger-style training process has not been explored previously.

\textbf{Simulation-to-Real Transfer.} Because we train policies in simulation and then evaluate on real robots, our work also relates to the literature on sim-to-real transfer~\cite{DBLP:conf/rss/SadeghiL17,DBLP:conf/icra/ByravanHHBNHMBSVH23, DBLP:conf/corl/RudinHR021, DBLP:conf/rss/0002FMM24, DBLP:journals/ral/XieCCQXSXWS23, gong2023arnold, akkaya2019solving, DBLP:conf/corl/ZhangWSWZT23, DBLP:conf/corl/0001WZ0024, DBLP:journals/corr/abs-2411-02189,DBLP:conf/corl/LumLCSASB24, DBLP:conf/rss/TangLAHS0FN23, DBLP:conf/icra/SuJQZMH024}. Domain randomization techniques~\cite{tobin2017domain,sadeghi2017cad2rl, zeng2024poliformer, handa2023dextreme} have proven effective for learning robust policies that transfer to the real world by training on sufficiently varied simulated observations. More recently, Singh et al.~\cite{singh2023splatsim} demonstrated zero-shot transfer of a high-dimensional visuomotor policy (for dexterous grasping) from photorealistic simulation to a real robot hand. In the context of multitask learning, Tang et al.~\cite{tang2023assembly} report successful zero-shot sim-to-real deployment of both specialist and generalist assembly policies. Inspired by these advances, we show that our generalist policy—trained with data-efficient DAgger in simulation—can likewise perform real-world tasks with minimal degradation, without any real-world fine-tuning.

\section{Method}
\label{sec:method}
We aim to train a single \textbf{multitask visuomotor policy} $\pi_G$ capable of performing $N$ distinct manipulation tasks, by efficiently distilling knowledge from $N$ pre-trained task-specific expert policies $\{\pi_1^*, \dots, \pi_N^*\}$. Our core contribution is a data-efficient multitask DAgger algorithm that adaptively allocates expert queries based on the generalist's learning progress and current performance across tasks. Below, we outline the problem setup and then detail our algorithm framework.

\subsection{Problem Formulation}
We model each task $i \in \{1,\dots,N\}$ as a Markov Decision Process (MDP) $\mathcal{M}_i = (\mathcal{S}_i, \mathcal{A}, P_i, r_i, \gamma)$, where $\mathcal{S}_i$ is the state space, $\mathcal{A}$ is the shared action space, $P_i$ is the transition dynamics, $r_i$ is the reward function, and $\gamma$ is the discount factor. We use reinforcement learning to learn a expert policy $\pi^*_i(a\,|\,s)$ for each task, typically operating on low-dimensional state $s \in \mathcal{S}_i$. 

Our goal is to train a single multitask policy $\pi_G(a\,|\,o, z_i)$ that maps high-dimensional observations $o \in \mathcal{O}$ (e.g., point clouds or images) and a task identifier $z_i$ (e.g., a one-hot or learned task embedding) to actions $a \in \mathcal{A}$. The multitask policy is trained via imitation learning to mimic the experts’ behavior. Standard DAgger~\cite{ross2011dagger} iteratively collects expert labels $a^* = \pi^*_i(s)$ for states $s$ encountered when rolling out the current learner policy $\pi_G^{(k-1)}$ for each round $k$. The policy is then retrained on the aggregated dataset $D^{(k)} = D^{(k-1)} \cup \{\text{new relabeled data}\}$. In the multitask setting, a key challenge is determining how to distribute the budget for collecting new expert demonstrations $\{(o, a^*, i)\}$ across the $N$ tasks at each iteration $k$. 
Naive uniform allocation can be inefficient: it may spend demonstrations on tasks that the policy already handles well, while failing to provide enough data on tasks where the policy struggles. Therefore, we develop a performance-aware scheduling strategy (Section~\ref{subsec:scheduler}) to intelligently allocate queries to the tasks that need them most.

\subsection{Single-task State-Based Expert Policy Training} 
\label{subsec:expert_training} 
Before starting the multitask DAgger pipeline, we first train individual expert policies for each task.  Each expert is trained with reinforcement learning algorithms like Proximal Policy Optimization (PPO)~\cite{schulman2017ppo}. A key aspect of this stage is that these single-task experts operate on low-dimensional state representations of the environment (e.g., proprioceptive information, object poses). Training RL policies using state observations is generally easier than using visual observations and leads to robust expert performance for individual tasks. These state-based experts then provide demonstration data for our primary multitask policy, which, in contrast, is trained to operate directly from high-dimensional sensory inputs (e.g., point clouds or RGB images). This strategy effectively circumvents the challenges of learning complex visuomotor skills from scratch using RL by leveraging the guidance of easily trained state-based experts within an imitation learning framework.
\subsection{Data-Efficient Multitask DAgger}
\label{subsec:multitask-dagger}

Our approach extends the classic Dataset Aggregation (DAgger) algorithm~\cite{ross2011dagger} to the multitask setting by incorporating a performance-aware scheduler for data collection. We maintain a single vision-based \emph{multitask} policy $\pi_G$ that is trained to perform all $N$ tasks using demonstrations from task-specific \emph{expert} policies $\{\pi^*_i\}_{i=1}^N$. The training process is iterative and interleaves policy improvement with targeted data aggregation (summarized in Algorithm~\ref{alg:multitask-dagger}). We begin by training an initial policy $\pi_G^{(0)}$ on a small dataset $D^{(0)}$ of expert demonstrations using behavioral cloning (BC) via supervised imitation loss minimization:
\begin{equation}
\small
\label{eq:bc-loss-generalist1}
    \pi_G^{(0)} = \arg\min_{\pi_G} \sum_{i=1}^N \sum_{(o, a^*, z_i) \in D_i^{(0)}} \mathcal{L}_{\text{BC}}\bigl(\pi_G(a \mid o, z_i),\, a^*\bigr),
\end{equation}
where $\mathcal{L}_{\text{BC}}$ is the imitation loss (e.g., Mean Squared Error for continuous actions). This yields a reasonable starting policy. We also set an initial \emph{DAgger mixing coefficient} $\epsilon_0 \in [0,1]$, controlling the probability of executing the learned policy versus the expert during data collection; $\epsilon_0$ can be initially small to favor expert actions for safety.

The core of the algorithm iterates for $K$ rounds ($k=1, \dots, K$). In each round $k$, the current multitask policy $\pi_G^{(k)}$ is first refined by training on the aggregated dataset $D^{(k-1)}$ using the BC loss (Eq.~\ref{eq:bc-loss-generalist1}). Then, the performance-aware scheduler (detailed in Sec.~\ref{subsec:scheduler}) analyzes task-specific performance metrics (like loss reduction or success rate) to determine an allocation budget $n_k^i$ for collecting new demonstrations for each task $i$, prioritizing tasks where $\pi_G^{(k)}$ underperforms or shows high learning potential, such that the total demonstrations collected $\sum_i n_k^i$ approximates the round budget $B$. Following the schedule, new data is collected by deploying $\pi_G^{(k)}$ for $n_k^i$ episodes per task. During these rollouts, actions are chosen by mixing the policy's output and the expert's action based on the previous mixing coefficient $\epsilon_{k-1}$. Crucially, the expert's action $a_t^*$ is always recorded alongside the observation $o_t$ and task identifier $z_i$. This newly collected data $\Delta D_k^i = \{(o_t, a_t^*, z_i)\}$ for all tasks is aggregated into the main dataset: $D^{(k)} = D^{(k-1)} \cup \bigcup_{i=1}^N \Delta D_k^i$. Finally, the mixing coefficient is decayed (e.g., $\epsilon_k = \max(\epsilon_{k-1} - \Delta \epsilon, \epsilon_{\min})$), gradually shifting control to the learned policy. This iterative process of refining the policy, scheduling data collection based on performance, collecting targeted demonstrations, and aggregating data continues until a stopping criterion is met, resulting in the final policy $\pi_G^{(K)}$.

\begin{figure}[t]
    \begin{small}
    \centering
    \fbox{ 
    \begin{minipage}{0.95\linewidth} 
    \textbf{Algorithm 1: Data-Efficient Multitask DAgger}\\[4pt] 
    \textbf{Require}: $N$ tasks, experts $\pi_i^*$, initial demos $D^{(0)}$, iterations $K$, budget $B$ per iteration, initial epsilon $\epsilon_0$, min demos $n_{\min}$, epsilon decay $\Delta\epsilon$. \\
    Train $\pi_G^{(0)}$ on $D^{(0)}$ (Eq.~\ref{eq:bc-loss-generalist1}). Initialize KF states $\{\hat{p}^i_0, P^i_0\}_{i=1}^N$. \\
    \textbf{for} ~$k=1$ to $K$:
    \begin{algorithmic}[1] 
    
        \State \textbf{Train Policy:} Update $\pi_G^{(k)}$ by training on $D^{(k-1)}$. Record task losses $L^i_{k-1, \text{start}}, L^i_{k-1, \text{end}}$. Compute gain $g^i_{k-1} = \max(0, L^i_{k-1, \text{start}} - L^i_{k-1, \text{end}})$.
        
        \State \textbf{Schedule Data Allocation:}
            \begin{itemize} \itemsep0pt 
                \item Update Kalman estimates $\hat{p}^i_{k-1}$ using measurements $\bar{p}^i_{k-1}$ from previous DAgger collection (iteration $k-1$) via standard Kalman filter update.
                \item Compute normalized need $\tilde{d}^i_{k-1}$ (based on $1-\hat{p}^i_{k-1}$) and gain $\tilde{g}^i_{k-1}$ (based on $g^i_{k-1}$) using rank normalization.
                \item Calculate priority score $\text{score}^i_{k-1} = \tilde{g}^i_{k-1}$ or $\tilde{d}^i_{k-1}$ and allocate $n_k^i$ demos per task using softmax over scores.
            \end{itemize}
            
        \State \textbf{DAgger Data Collection:} For each task $i$, collect $n_k^i$ trajectories using $\pi_G^{(k)}$ mixed with $\pi_i^*$ (with prob $\epsilon_{k-1}$), storing expert labels $(o_t, a_t^*, z_i)$ in $\Delta D_k^i$. Measure raw success rate $\bar{p}_k^i$ for next iteration's scheduling.
        
        \State \textbf{Update Dataset and Parameters:} $D^{(k)} \leftarrow D^{(k-1)} \cup \bigcup_i \Delta D_k^i$. Decay $\epsilon_k \leftarrow \max(\epsilon_{k-1} - \Delta \epsilon, \epsilon_{\min})$.
    \end{algorithmic}
    \textbf{end for}\\
    \textbf{Return} $\pi_G^{(K)}$
    \end{minipage}
    }
    \caption{Overview of the proposed data-efficient multitask DAgger algorithm.}
    \label{alg:multitask-dagger}
    \end{small}
\end{figure}


\subsection{Performance-Aware Scheduling via Adaptive Filtering and Allocation}
\label{subsec:scheduler}
The scheduler determines the allocation $\{n_k^i\}_{i=1}^N$ of expert demonstrations for the current iteration $k$ based on performance metrics observed up to the previous iteration ($k-1$). The core idea is to prioritize tasks based on either the estimated 'Task Need' or the measured 'Performance Gain'. We evaluate these two criteria separately:

\paragraph{Task Need (Inverse Success Rate):} We quantify how much a task currently needs improvement by considering the policy's success probability on it. A lower success probability signifies a higher need for corrective expert data. Specifically, we track the filtered success probability estimate $\hat{p}^i_{k-1}$ (detailed below) and use the \emph{inverse success probability}, $1 - \hat{p}^i_{k-1}$, as the metric for measuring TN. The motivation is straightforward: allocate more resources to tasks where the policy is frequently failing, as these represent the largest gaps in performance. 


\paragraph{Performance Gain (Learning Progress):} Alternatively, we consider how rapidly the policy is improving on each task, measured by the magnitude of the drop in imitation loss within one DAgger iteration.  The PG for task $i$ from DAgger iteration $k-1$ is $g^i_{k-1} = \max(0, L^i_{k-1, \text{start}} - L^i_{k-1, \text{end}})$, where $L^i_{k-1, \text{start}}$ describes the loss at the beginning of training for this iteration $k-1$, and $L^i_{k-1, \text{end}}$ describes the loss after a predefined number of training steps. Large PG suggests the task is in a "steep" region of the learning curve, and additional data might yield substantial further improvements efficiently. The motivation here is to capitalize on learning momentum: invest more data where the policy is currently learning effectively, potentially accelerating convergence on those tasks. PG can be particularly useful as an alternative scheduling criterion when initial success rates are near zero for multiple tasks, making TN less discriminative for prioritizing data collection.

To make robust scheduling decisions, we need a stable estimate of each task's success probability. Raw success rates measured during data collection can be noisy, especially when a task is allocated only a few demonstration rollouts in a given round. A short string of lucky or unlucky outcomes can give a misleading impression of the policy's true capability on a task. To filter out this noise, we employ a Kalman filter for each task. The filter treats the true success probability as a hidden state and recursively updates its belief by combining the previous estimate with the new, noisy measurement. This provides a principled way to weigh new evidence against our prior belief, trusting measurements from many rollouts more than those from few. This smoothed estimate, $\hat{p}^i$, provides a more reliable signal for our 'Task Need' (TN) scheduling metric.

Empirically, we found that directly using the two metrics without normalization can lead to some overly tilted data distribution. For example, if one hard task got 0 success rate and the others got over 50\% success rate, almost all the data are collected from that hard task for the next iteration. To ensure robustness regardless of the chosen scheduling metric ('need' derived from $1-\hat{p}^i_{k-1}$ or 'gain' $g^i_{k-1}$), which may have different scales and distributions, we employ rank normalization. For the selected metric $m$ (either need or gain), let $\{m^1, \dots, m^N\}$ be the values across the $N$ tasks. Let $\text{rank}(m^i)$ denote the rank of task $i$ based on this metric (e.g., rank 1 for lowest value, rank $N$ for highest value). The rank-normalized value $\tilde{m}^i$ is computed as:
\begin{equation}
\label{eq:rank_norm}
    \tilde{m}^i = \frac{\text{rank}(m^i) - 1}{N - 1}, \quad \text{for } N > 1 \quad (\text{and } \tilde{m}^i = 0 \text{ if } N=1).
\end{equation}
This scales the ranks to the range $[0, 1]$, preserving the relative ordering while making the metrics scale-invariant and robust to outliers. We compute the normalized need $\tilde{d}^i_{k-1}$ (ranking $1-\hat{p}^i_{k-1}$) and normalized gain $\tilde{g}^i_{k-1}$ (ranking $g^i_{k-1}$) using this method.

We then take the outputs from normalization and use them as a priority score for each task $i$. We convert these scores into allocation probabilities using softmax and decide the number of trajectories allocated for each task by taking the product of this probability and the budget for the total number of trajectories to be allocated at each round. This adaptive mechanism dynamically focuses expert queries on tasks that are most likely to benefit, driving data efficiency.


\section{Experiments}
\label{sec:experiments}
In this section, we evaluate our MultiTask DAgger algorithm, designed to enhance sample efficiency in multitask imitation learning. We conduct experiments in two distinct domains: the standard Meta-World benchmark, and a set of drawer-opening tasks in IsaacLab~\cite{mittal2023orbit}. Our goals are to demonstrate: 
(1) Improved sample efficiency and asymptotic performance compared to standard Behavior Cloning (BC) and uniform DAgger approaches.
(2) The effectiveness of our performance-aware scheduling mechanism.
(3) The effectiveness of the trained multitask policy in real-world hardware experiments through \textbf{zero-shot sim-to-real transfer}.
\begin{figure}[!htb]
    \centering
    \includegraphics[width=0.36\linewidth]{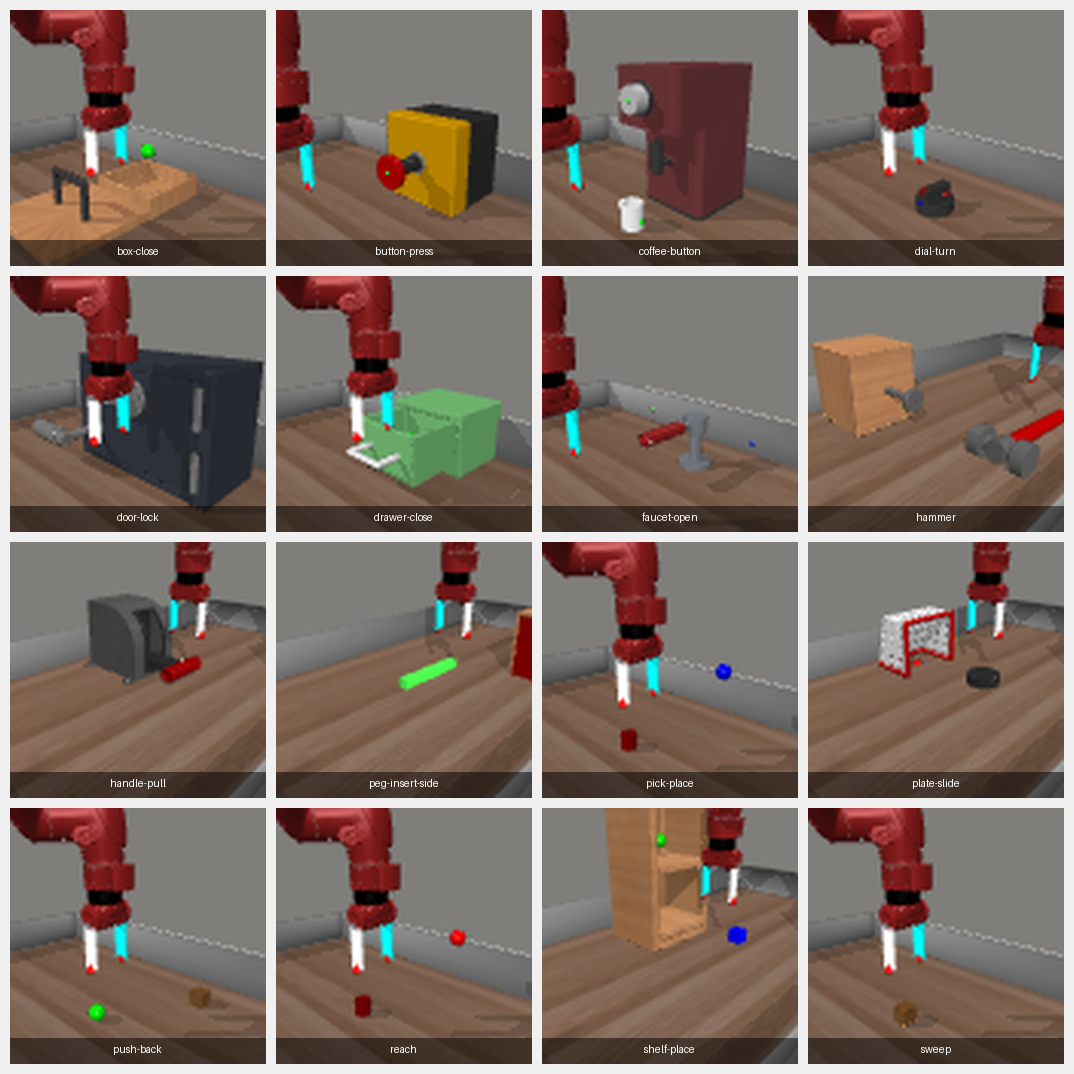}
    \includegraphics[width=0.6\linewidth]{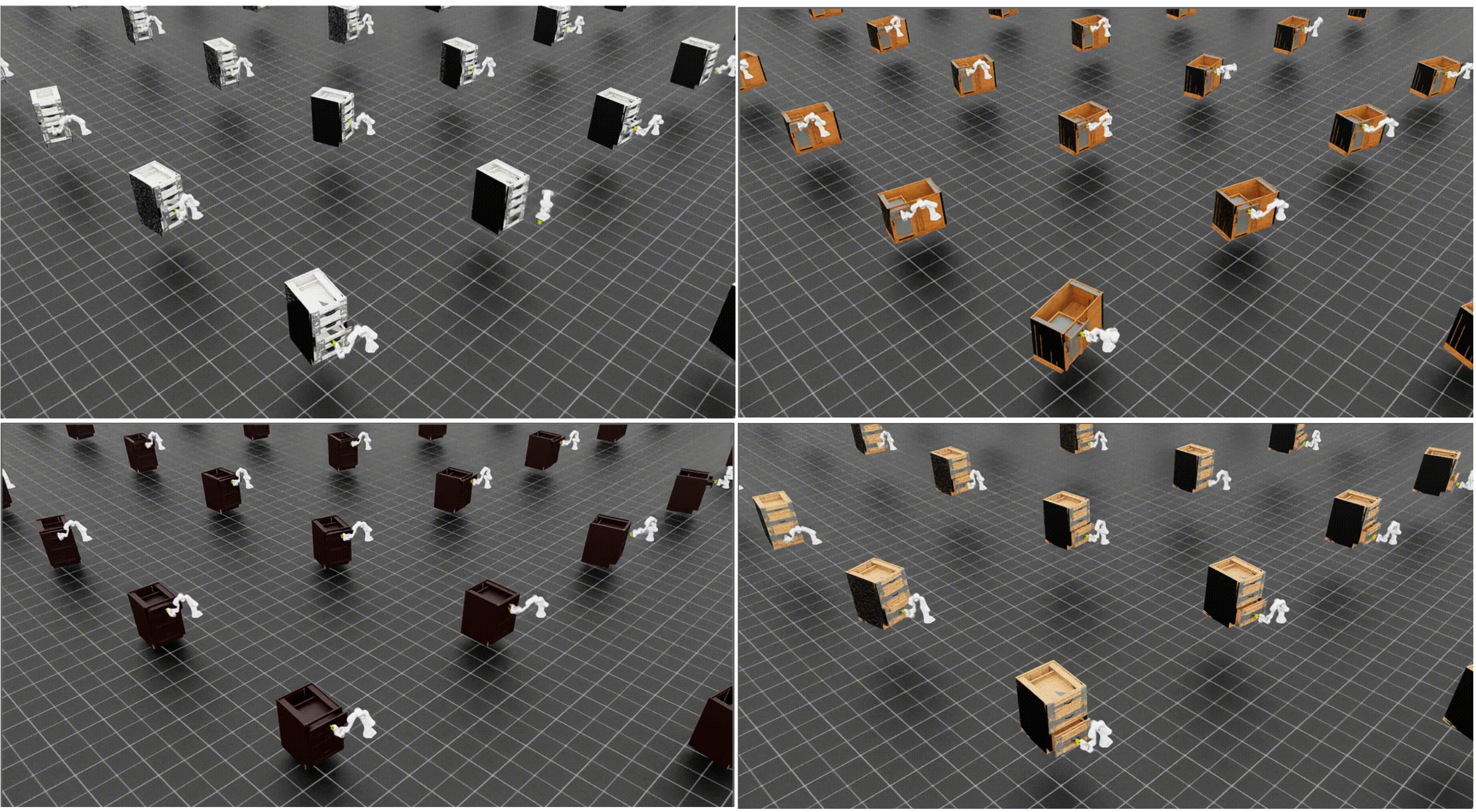}
    \caption{Examples of Meta-World tasks and IsaacLab Drawer tasks used in our experiments.}
    \label{fig:metaworld}
\end{figure}
\subsection{Meta-World Benchmark and Drawer-Opening in IsaacLab}
To evaluate scalability and performance on a standard benchmark, we utilize the MetaWorld~\cite{DBLP:conf/corl/YuQHJHFL19} suite, specifically the Multitask setup where we select 36 distinct manipulation tasks. We first train a state-based Soft-Actor-Critic (SAC)~\cite{DBLP:conf/icml/HaarnojaZAL18} expert policy for each task. Specifically, we apply the SAC algorithm to all 50 tasks comprising the benchmark, and success rates surpass 80\% on 36 of these tasks. Then we train multitask policies using our method under two conditions:
1.~\textbf{State-based Multitask Policy:} The multitask policy receives low-dimensional state observations, mirroring the expert's input space but requiring generalization across tasks.
2.~\textbf{Image-based Multitask Policy:} The multitask policy receives RGB image observations, requiring learning from high-dimensional visual input in addition to multitask generalization. This setup allows us to compare performance and sample efficiency against baselines under varying input modalities and task complexities.
\begin{figure*}[!htb]
    \centering
    \includegraphics[width=0.26\linewidth]{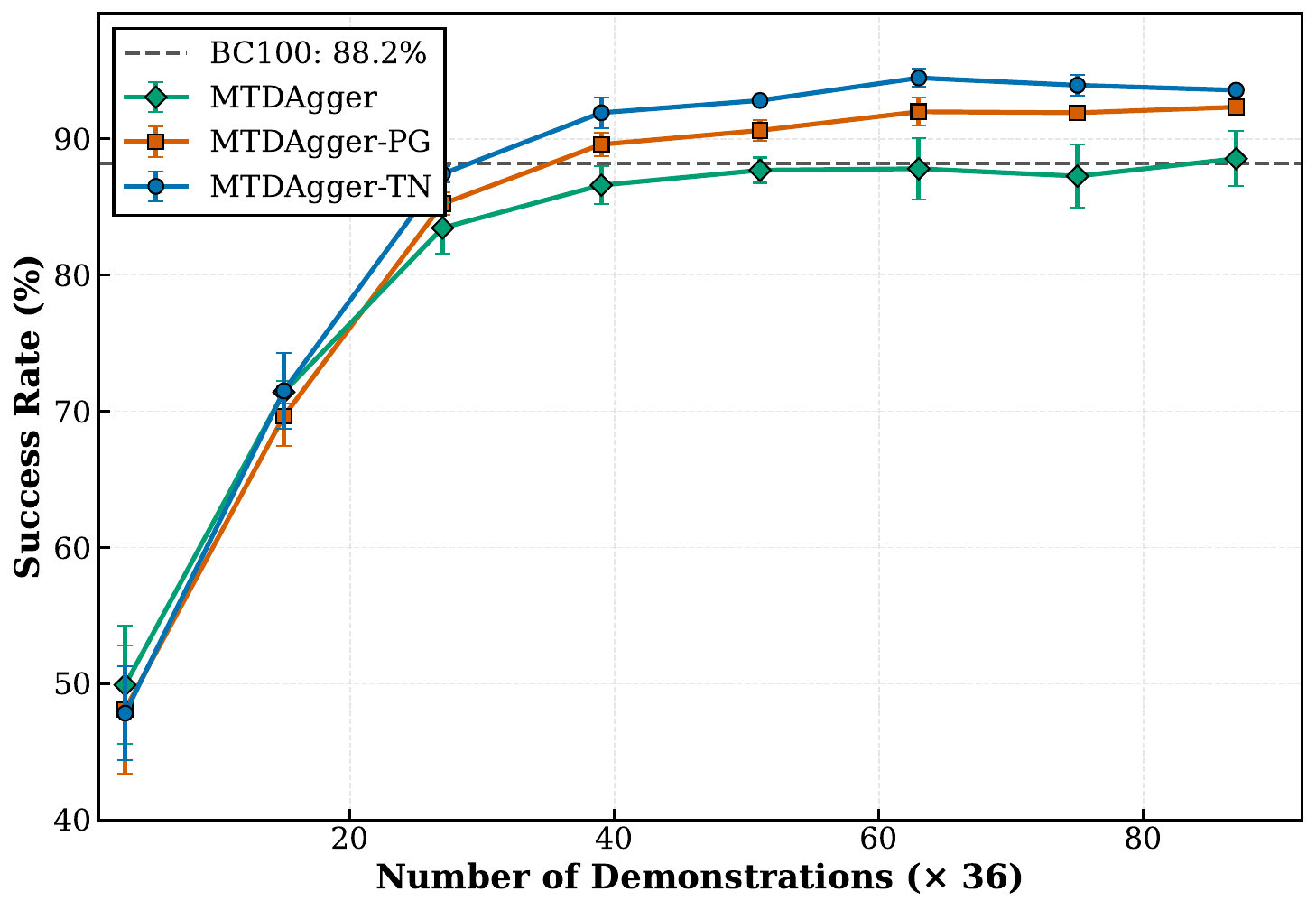}
    \includegraphics[width=0.26\linewidth]{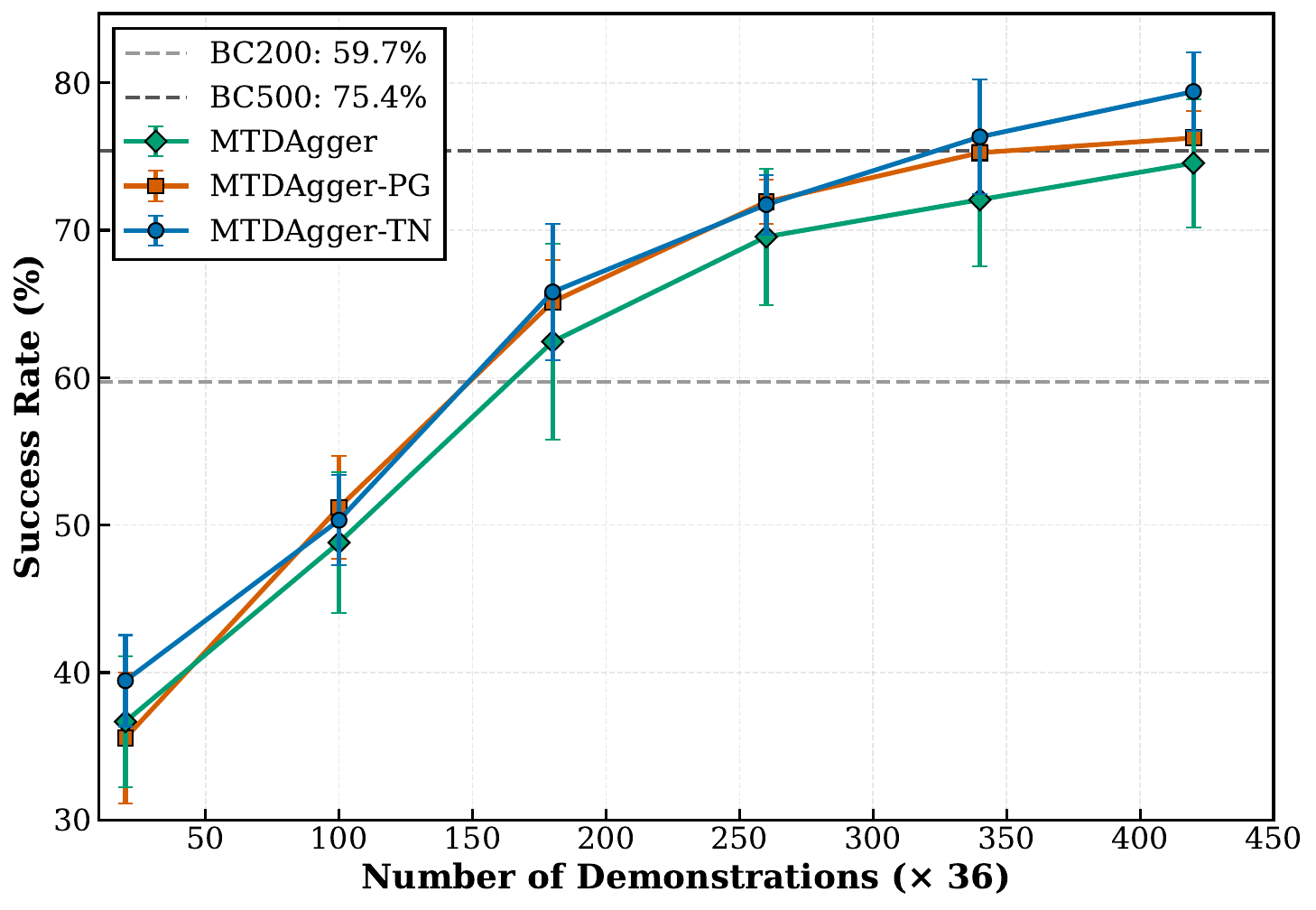}
        \includegraphics[width=0.26\linewidth]{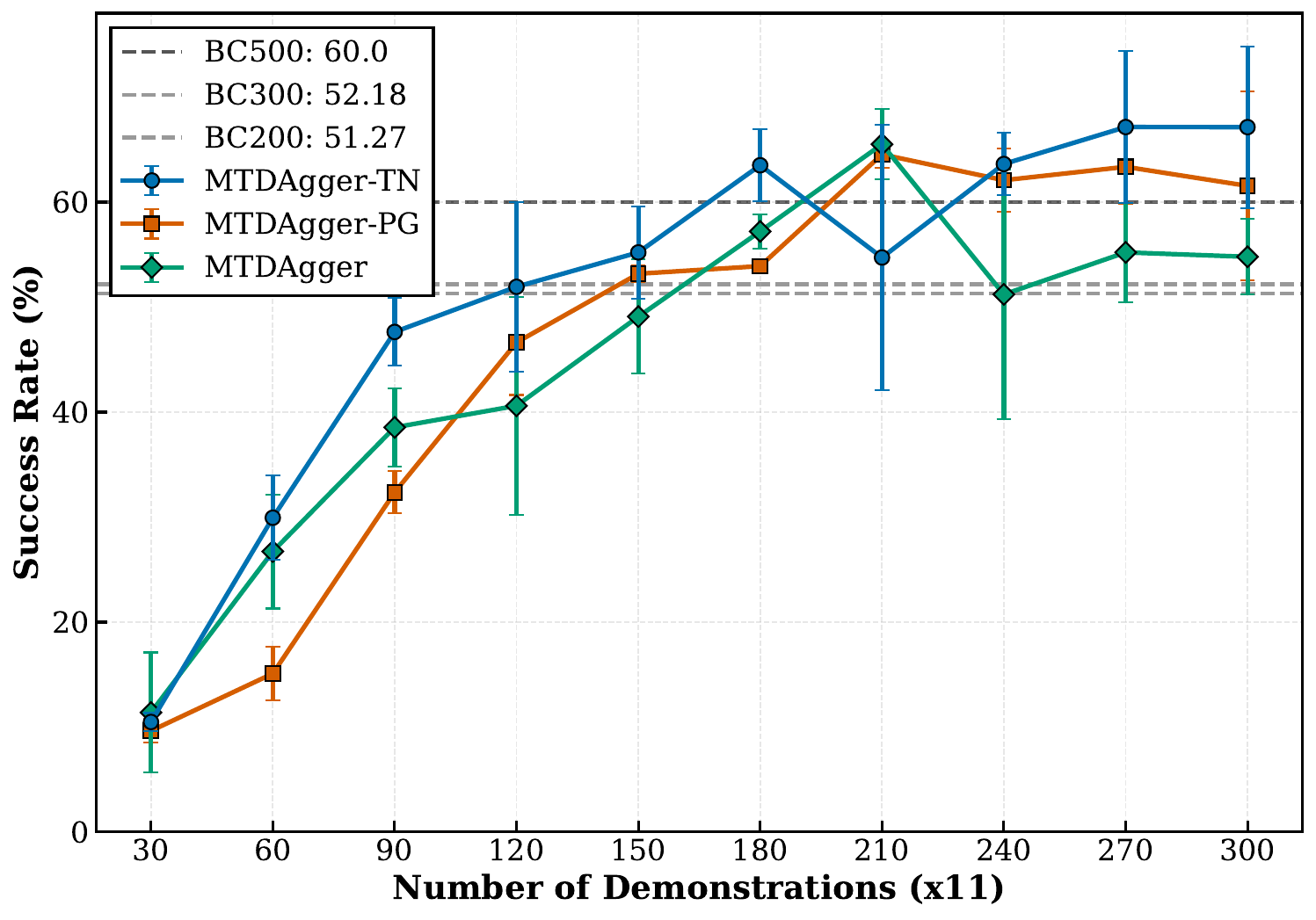}
    \caption{Average success rate  vs. the number of expert demonstrations collected per task (averaged). Left: Metaworld state-based multitask policy comparison. Middle: Metaworld pixel-based multitask policy comparison. Right: IsaacLab Drawer point-cloud-based multitask policy comparison. Our performance-aware scheduling mechanism generally achieves higher sample efficiency and final performance compared to standard BC with varying data budgets and Uniform DAgger.}
    \label{fig:learning}
\end{figure*}
To assess generalization across geometrically diverse but functionally similar objects, we use an IsaacLab environment featuring a Franka robot arm. The task involves opening various drawers. The training task suite includes 11 different drawers. 10 of them are sourced from the PartNet-Mobility~\cite{Mo_2019_CVPR, Xiang_2020_SAPIEN} dataset, selected for their diverse geometries and articulation mechanisms, plus one which models a real drawer that we have. Each task requires learning a policy to successfully open its specific drawer. Single-task expert policies are trained using PPO~\cite{schulman2017ppo} on low-dimensional state representations. Our goal is to train a single, multitask policy using our adaptive DAgger approach that operates directly from high-dimensional camera observations and can successfully operate all 11 drawers. A visualization of the simulated setup with the standard drawer is shown in Figure~\ref{fig:metaworld} right.
\begin{figure}[!htb]
    \centering
    
    \includegraphics[width=1\linewidth]{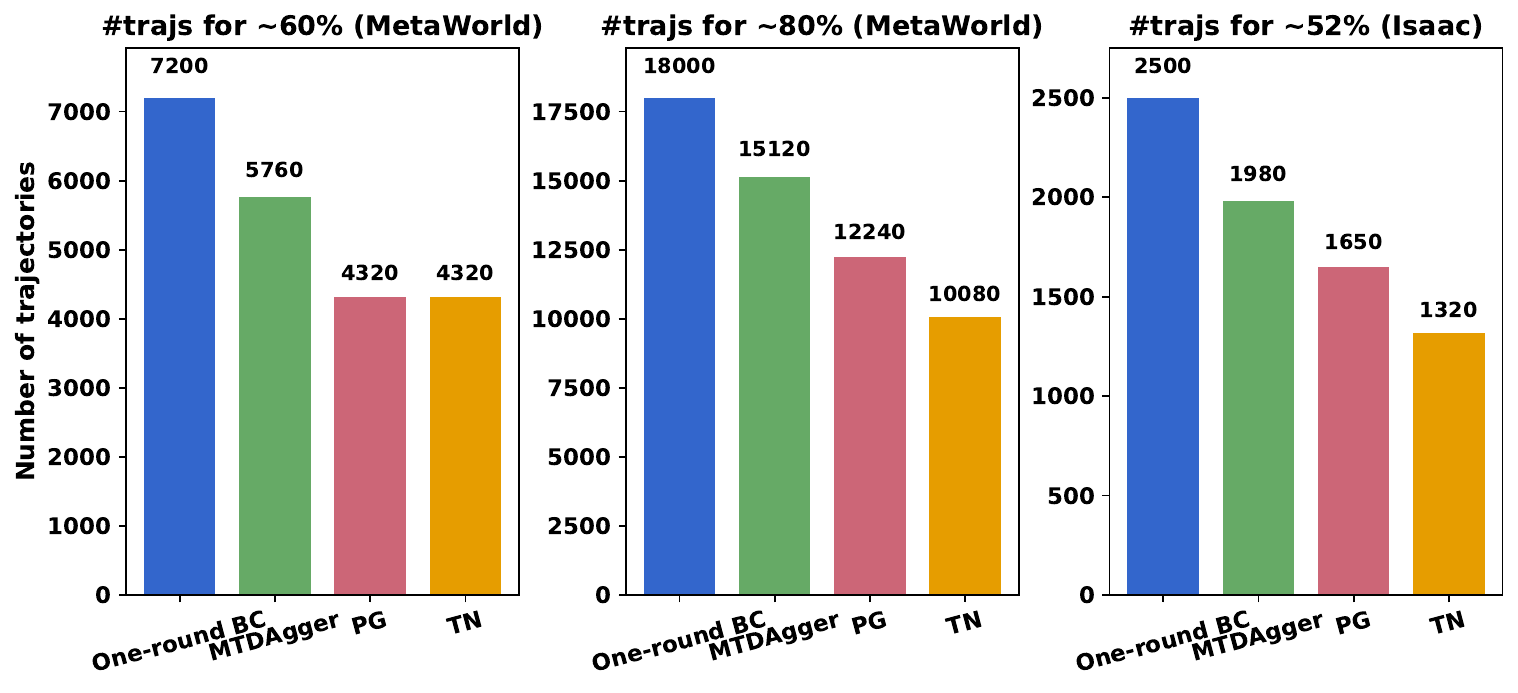}
    \caption{Comparison of number of trajectories needed to reach high success rates in MetaWorld and IsaacLab Drawer tasks.}
    \label{fig:chart}
\end{figure}
\subsection{Setup}
For the Meta-World state-based experiments, for each DAgger round, we collect 108 demonstrations in total per round across 36 tasks, starting with 3 initial demonstrations per task. For the Meta-World pixel-based experiments, for each round, we collect 720 demonstrations in total per round across 36 tasks, starting with 40 initial demonstrations per task. For the Isaac Drawer experiments, we ran for 10 rounds, collecting 30 demonstrations per task per round across 11 tasks. 

We compare our Data-efficient Multitask DAgger against two primary baselines: 1.~\textbf{BC} A standard imitation learning approach where a policy is trained offline on a fixed dataset of expert demonstrations collected uniformly across all tasks. 2.~\textbf{Uniform DAgger:} The standard DAgger algorithm where, in each round, an equal number of demonstrations are collected for every task, regardless of estimated difficulty or learning progress. The primary metric for evaluation is the average success rate across all tasks in the respective benchmark. Success rates are measured by deploying the current multitask policy checkpoint for a fixed number of episodes per task and calculating the percentage of successful completions. We report these average success rates throughout the training process (across DAgger rounds) to analyze sample efficiency.
\begin{table*}[htbp]
    \centering
    \small
    \caption{Sim-to-Real Performance on in-domain vs.\ out-of-domain tasks. (15 trials each)}

    \begin{tabular}{lcccc}
        \toprule
        \textbf{Method} & \textbf{In-Domain (Success Rates)} & \textbf{Unseen (Success Rates)} & \textbf{In-Domain (Task Progress)} & \textbf{Unseen (Task Progress)}\\
        \midrule
        Ours   & $\mathbf{0.60 \pm 0.28}$ & $\mathbf{0.53 \pm 0.25}$ & $\mathbf{2.40 \pm 0.57}$ & $\mathbf{2.27 \pm 0.75}$\\
        MTDAgger & $0.33 \pm 0.25$ & $0.13 \pm 0.19$ & $1.80 \pm 0.57$ & $0.67 \pm 0.52$\\
        BC     & $0.40 \pm 0.33$ & $0.13 \pm 0.09$ & $1.20 \pm 0.98$ & $0.40 \pm 0.28$\\
        \bottomrule
    \end{tabular}
\end{table*}
\subsection{Simulation Learning Performance Comparison}

Figure~\ref{fig:learning} presents the learning curves comparing the average success rates of our proposed multitask DAgger variants (MTDAgger-PG and MTDAgger-TN) against BC with varying data budgets and standard Uniform DAgger. Across all three experimental settings – MetaWorld state-based (left), MetaWorld pixel-based (middle), and IsaacLab point-cloud-based (right) – our performance-aware scheduling strategies consistently demonstrate superior sample efficiency and achieve higher final performance. Notably, MTDAgger-PG generally performs better than MTDAgger-TN, showing that the success rate is generally a more stable metric for calculating the task difficulty, even though it is calculated during data collection. This efficiency gain is further quantified in Figure~\ref{fig:chart}, which shows the substantial reduction in the total number of expert trajectories required by our methods to reach target success rates (e.g., ~60\% and ~80\% in MetaWorld, ~52\% in IsaacLab) compared to one-round BC and Uniform MTDAgger. These results validate the effectiveness of adaptively allocating data collection efforts based on TN or PG, leading to more data-efficient learning of multitask policies from multiple experts across diverse tasks and observation modalities.

\subsection{Analysis of Adaptive Scheduling}
\begin{figure*}[htbp]
    \centering
    \includegraphics[width=0.225\linewidth]{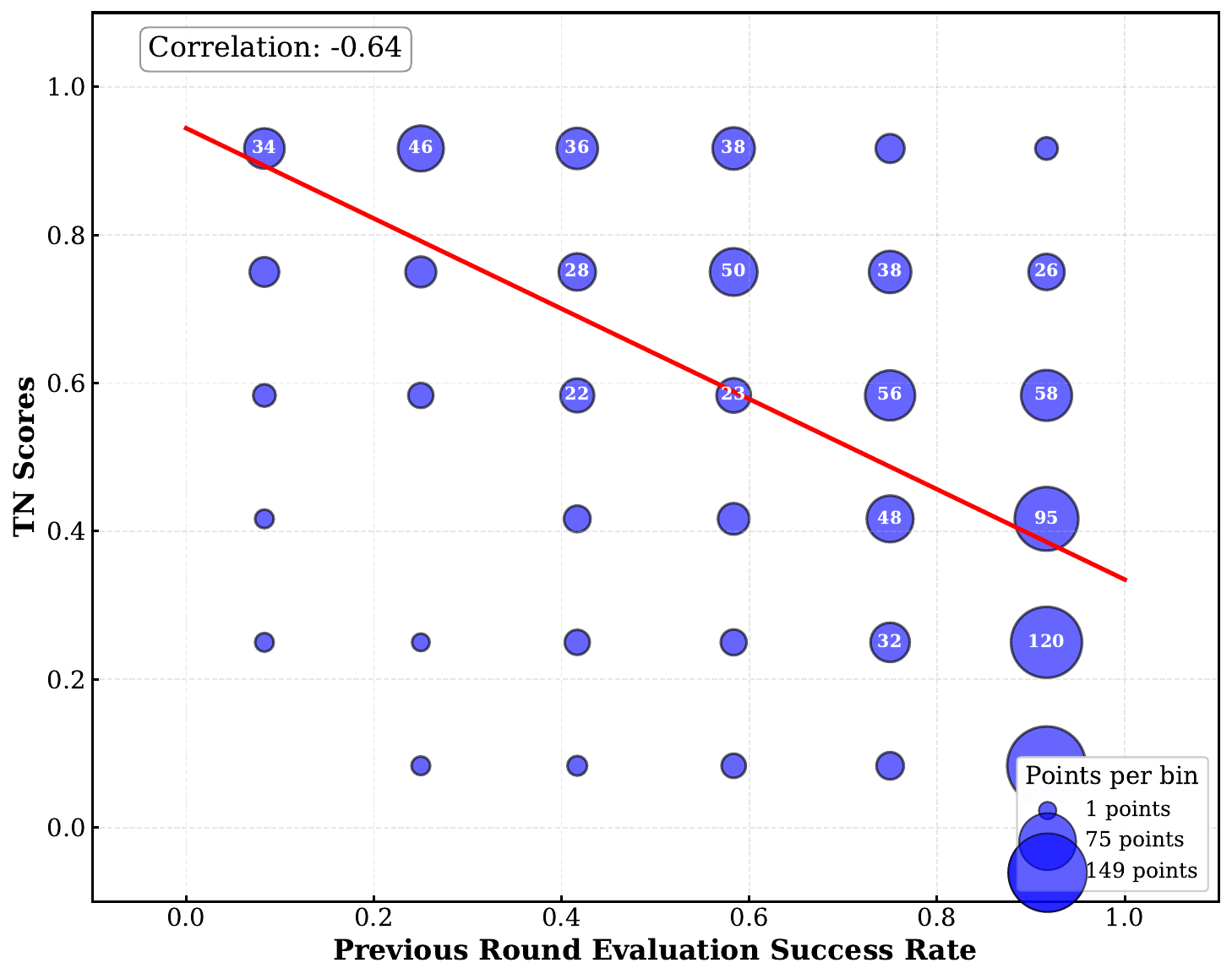}
    \includegraphics[width=0.225\linewidth]{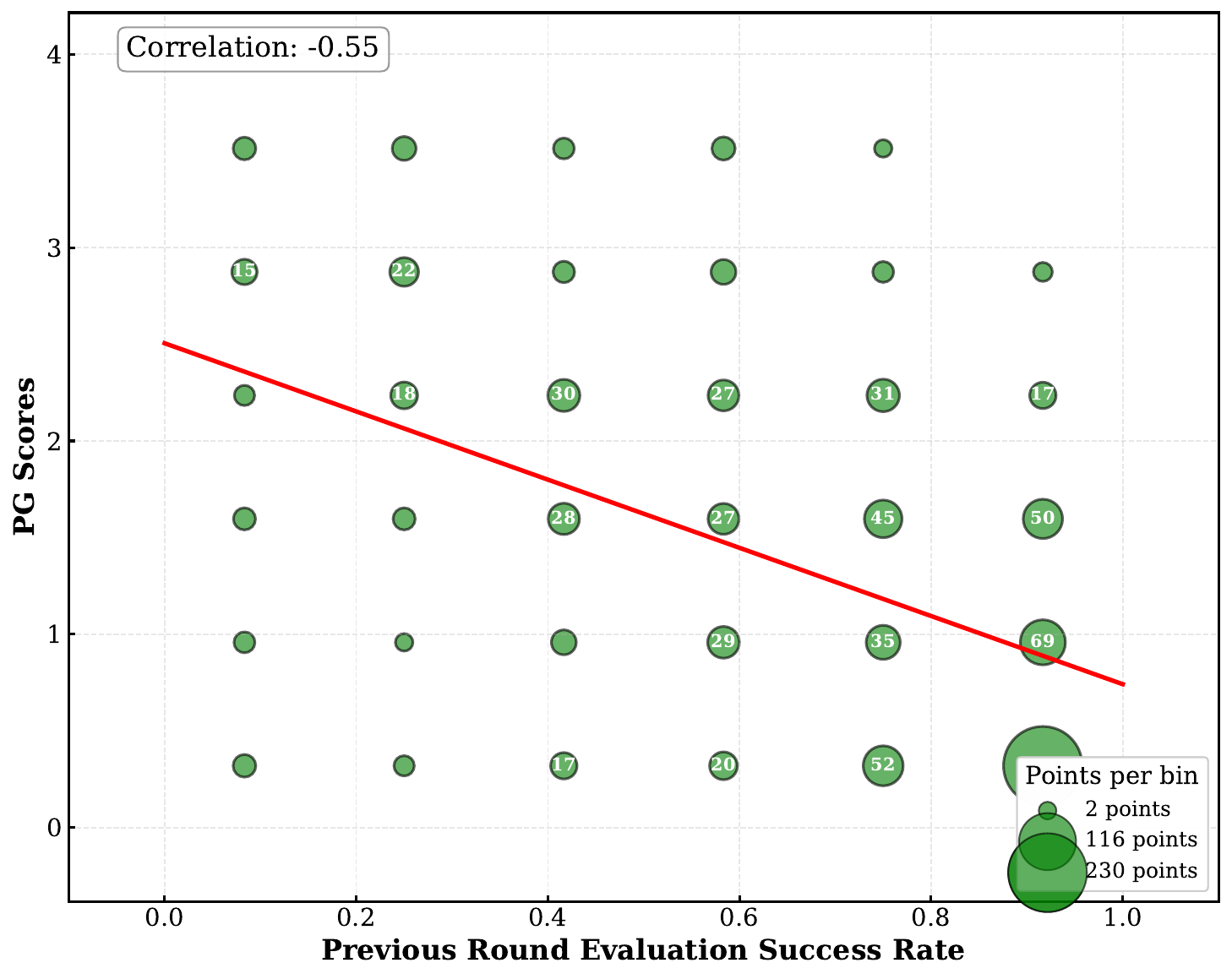}
        \includegraphics[width=0.26\linewidth]{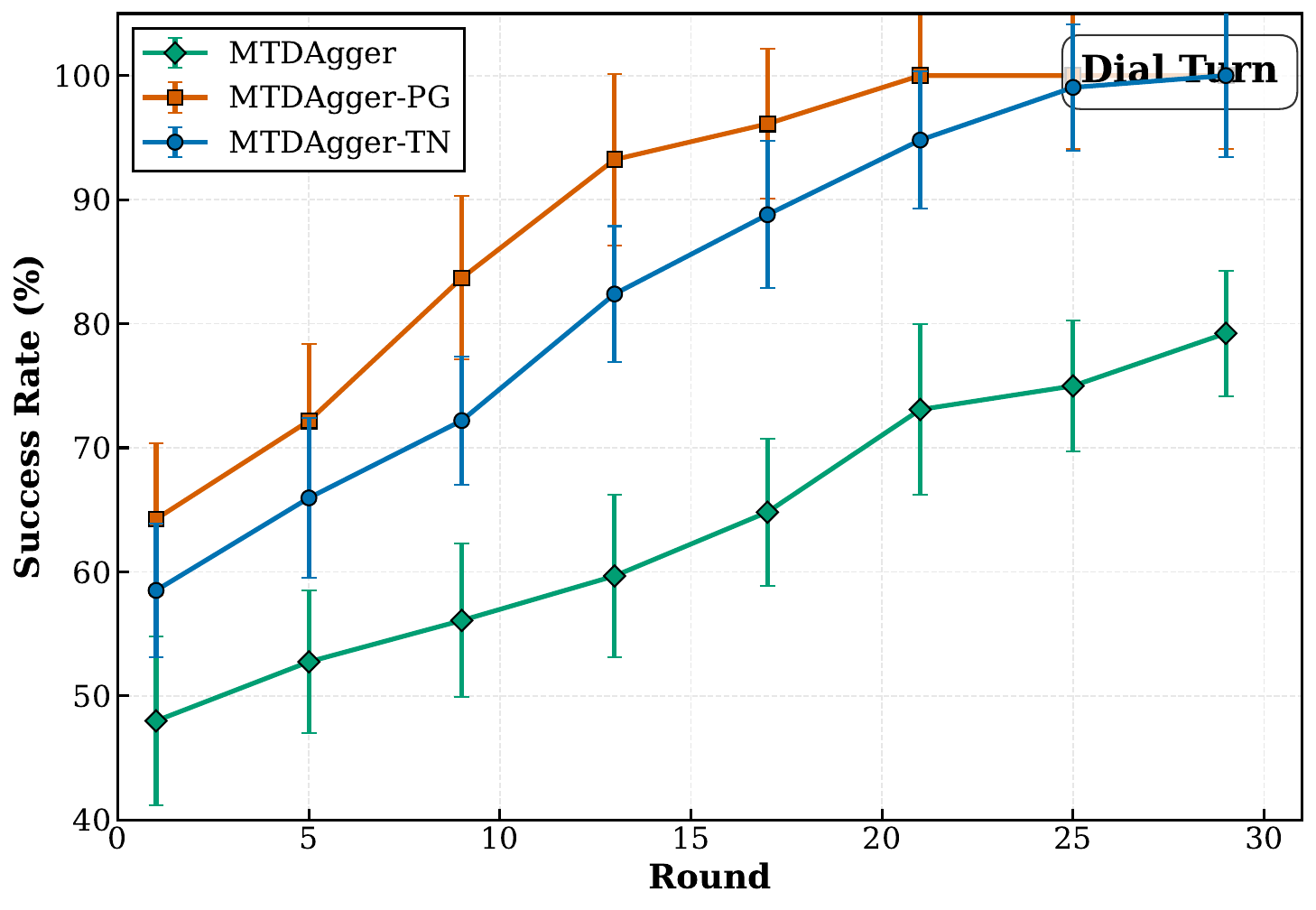}
    \includegraphics[width=0.26\linewidth]{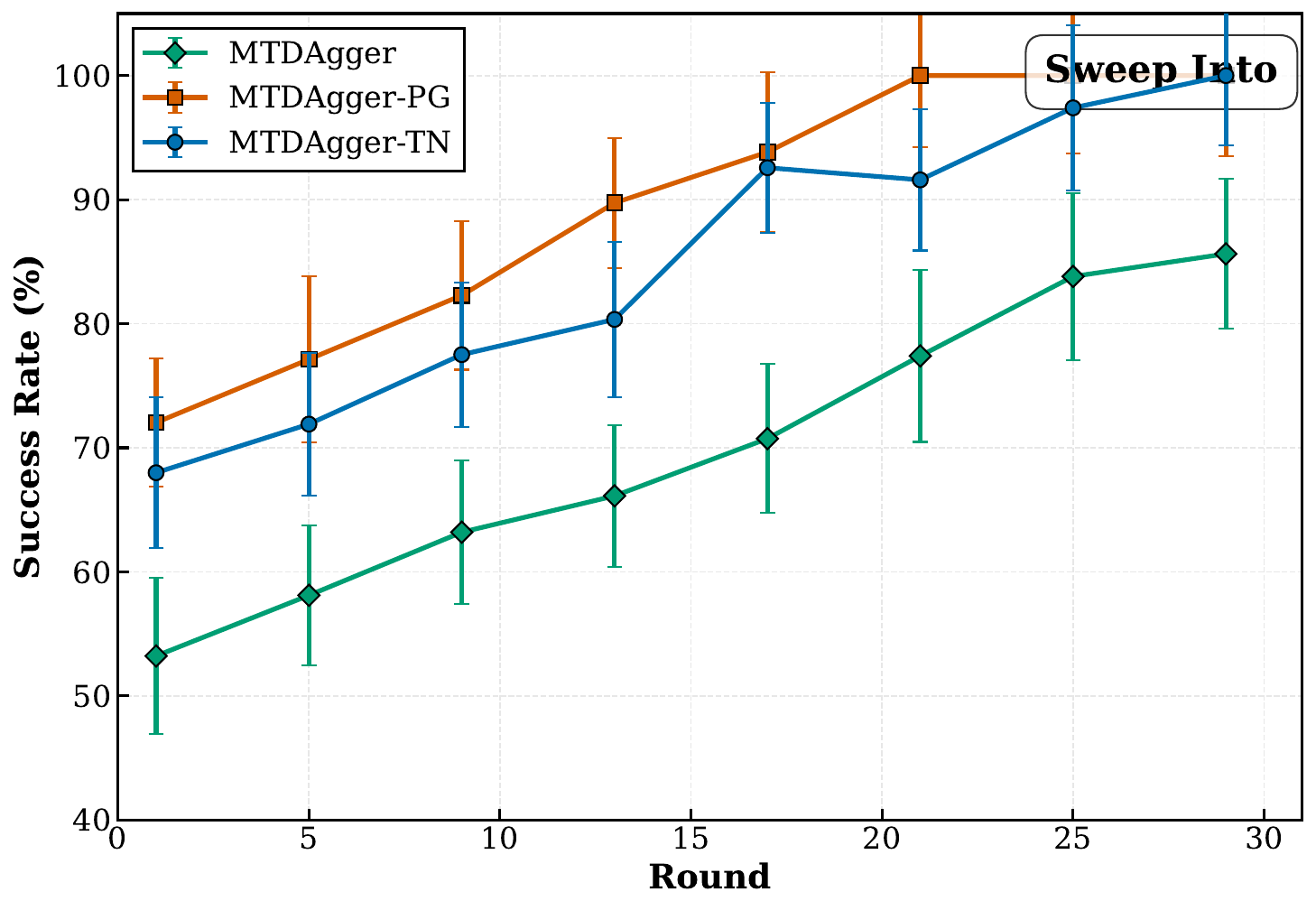}
    \caption{Left: correlation between the estimated TN metric (based on KF success rate estimate) or PG metric from the final training rounds and the final evaluation success rate across Meta-World tasks. Right: Success rate comparison on `dial-turn-v2` task and 'sweep-into-v2' task.}
    \label{fig:corr1}
\end{figure*}

A key component of our method is the ability to estimate task needs and allocate resources accordingly. We verify the effectiveness of our scheduling metrics. Figure~\ref{fig:corr1} first plot shows the correlation between the estimated TN (derived from the Kalman filter's success rate estimate) in the final rounds and the actual measured success rate during evaluation. Figure~\ref{fig:corr1} second plot similarly shows the correlation for the PG metric ($\Delta L_i$). The results show that for those tasks where the evaluation success rate is high, our metrics give it lower scores which means it will be scheduled to collect fewer trajectories next round, and vice versa. Both plots indicate a meaningful correlation, suggesting that our metrics effectively capture aspects of task difficulty and learning potential, thereby justifying their use in guiding the data allocation process. 

Figures~\ref{fig:corr1} right shows the learning curves for two of the most challenging tasks in the Meta-World setting, `dial-turn-v2` and `sweep-into-v2`, respectively. We run BC with 1000 demonstrations per task for all 36 tasks and find these two are
the hardest tasks (lowest success rates). Compared to the plots for the comparison of success rate averaged across all the tasks, we see on these hard tasks our methods demonstrate even better performance with a larger gap compared to Uniform DAgger. PG Metric shows better performance than TN Metric on these tasks, indicating that prioritizing tasks by learning progress can be especially effective for extremely challenging tasks.

\begin{figure}[!htb]
    \centering

    \begin{subfigure}[t]{0.44\linewidth}
        \centering
        \includegraphics[width=\linewidth]{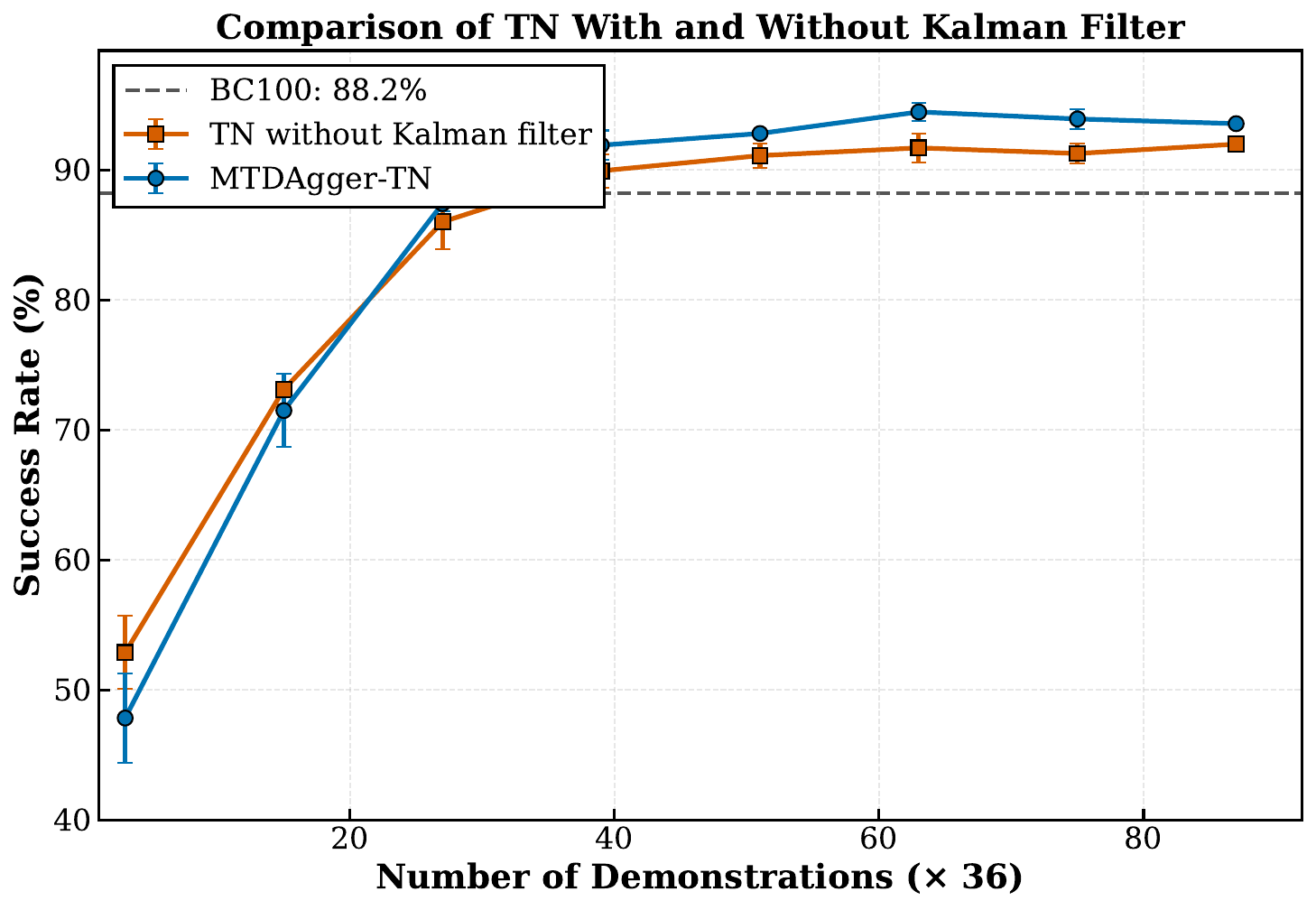}
        \caption{}  \label{fig:tn_comparison}
    \end{subfigure}
    \begin{subfigure}[t]{0.42\linewidth}
        \centering
        \includegraphics[width=\linewidth]{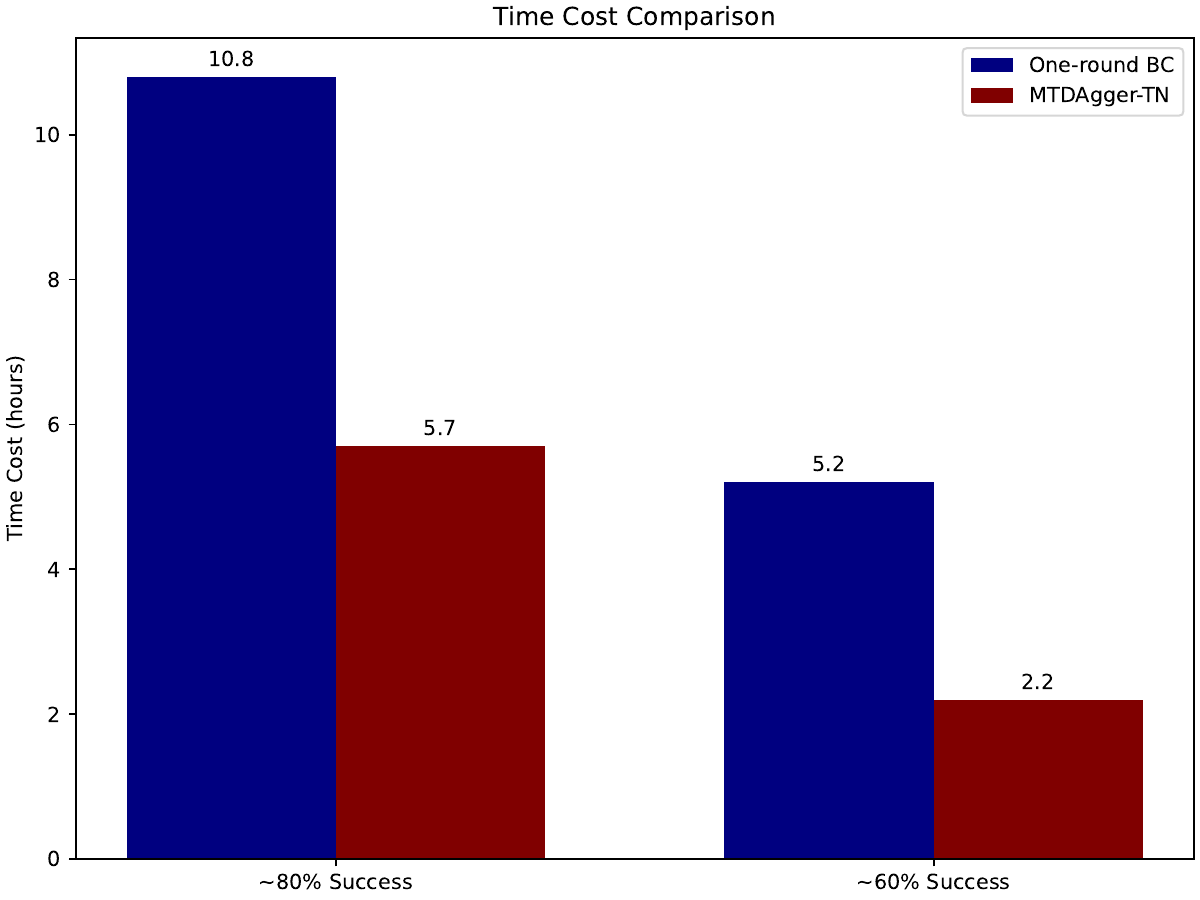}
        \caption{}  \label{fig:time_cost}
    \end{subfigure}

    \caption{Addtional experiment results. (a)  Performance if we do not use Kalman filter for MTDAgger-TN. (b) Wall time cost for training comparison on MetaWorld.}
    \label{exp:add}
\end{figure}

Figure~\ref{exp:add} presents supplementary results further analyzing the performance and characteristics of our proposed method. Figure~\ref{exp:add}(a) provides an ablation study on the MetaWorld state-based tasks, demonstrating that incorporating the Kalman filter (MTDAgger-TN) yields slightly better final performance and stability compared to using raw success rates for the Task Need scheduler (TN without Kalman filter), validating the benefit of smoothed success rate estimation. Figure~\ref{exp:add}(b) compares the wall-clock training time required to reach target success rates (60\% and 80\%) on MetaWorld pixel-based tasks, highlighting that MTDAgger-TN achieves these thresholds significantly faster than standard one-round Behavior Cloning, owing to its data efficiency. 
\begin{figure}[!htb]
    \centering
    \begin{subfigure}[b]{0.5\linewidth}
        \includegraphics[width=\linewidth]{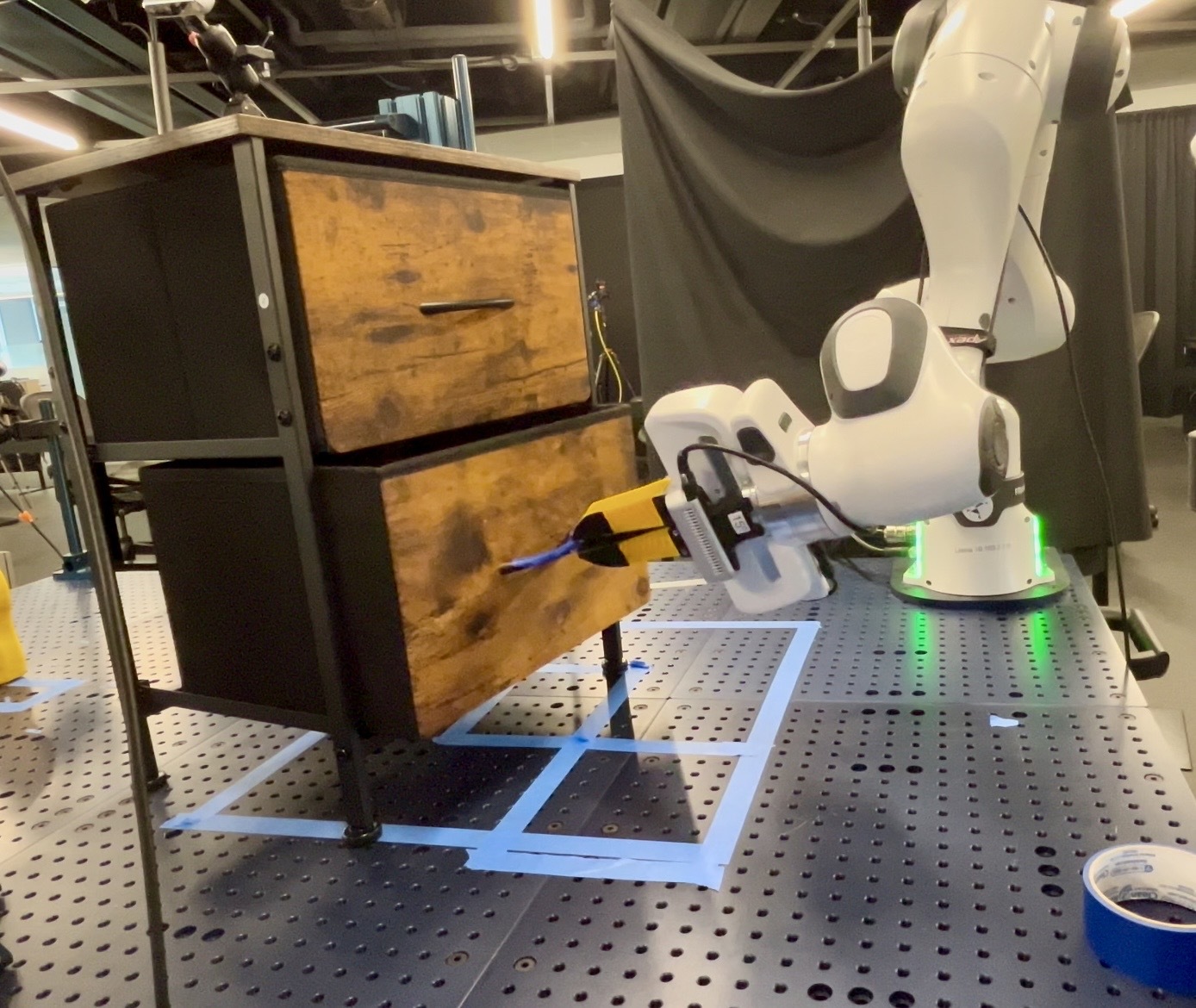}
        \caption{Unseen Drawer }
        \label{fig:real_drawer_a}
    \end{subfigure}
    \hfill
    \begin{subfigure}[b]{0.43\linewidth}
        \includegraphics[width=\linewidth]{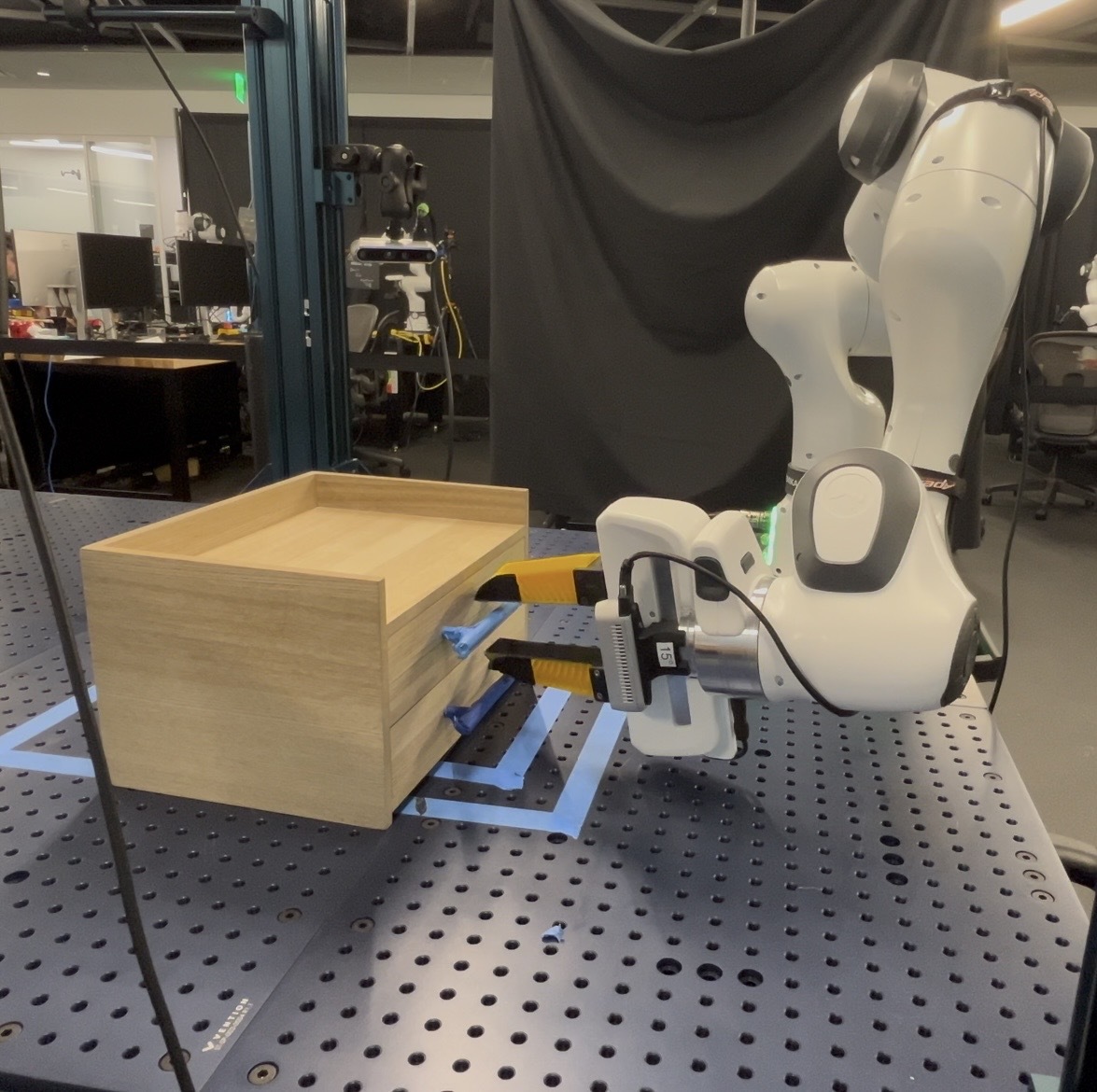}
        \caption{In-Domain Drawer }
        \label{fig:real_drawer_b}
    \end{subfigure}
    \caption{Images of the real-world Franka setup for the drawer opening task.}
    \label{fig:real_drawer}
\end{figure}
\subsection{Sim-to-Real Evaluation}
To assess the practical applicability of policies trained with our framework, we performed sim-to-real transfer experiments. For perception in the real world, the robot uses a Intel RealSense D435 camera. The raw point cloud data from the camera was cropped to remove background with 4096 points uniformly sampled. As shown in Figure~\ref{fig:real_drawer}, we tested on a domain drawer that is included in the simulation training set, as well as an unseen drawer that was not in the training data. We directly use the multitask policy we trained in simulation to rollout on the real robot, for three different drawer positions, 5 trials each. We measure the success rates on these two drawer for different methods, as well as the task progress, where the robot will get 1 point each for grasping the handle, attempting to pull, and opening the drawer fully (max 3 points). Due to the greater difficulty presented by the unseen drawer task, a less stringent success criterion was established, requiring only partial opening of the drawer for the attempt to be considered successful.  

\subsection{Robot Demos}
We show videos of real robot rollouts on in-domain and unseen drawers respectively\footnote{ \url{https://sites.google.com/brown.edu/mtdagger/home}}. It also contains 2 failure videos which still shows some interesting behavior the robot learned. In the \textbf{Failure1\_seen\_drawer} part, the robot was able to grasp the handle but then it slipped away when the gripper tried to pull the handle. But then the robot shows some interesting \textbf{recovery behavior} that it was able to find and grasp the handle again but just not further pull it our.
All the videos are shown in the \textbf{original speed} and we executed the learned policy on real robot with 10 Hz frequency.

\subsection{Other Metrics we have tried in the experiments}
Besides the two metrics -- Task Need and Performance Gain, that we have found to be useful in the experiments, in practice we have tried a few other metrics that didn't end up working well. These include:
\begin{itemize}
    \item KL divergence between the output distribution of the multitask policy and the expert policies
    \item Normalized final/initial loss of each task during each training round
    \item Variance of the normalized loss of each task during each training round
\end{itemize}

\section{Conclusion}
We presented a framework for efficiently training a multitask robot policy by integrating DAgger with a performance-aware task scheduler. Our approach significantly reduces the total amount of data required by prioritizing demonstrations for tasks that are either currently difficult or promise high learning gains. We showed that this strategy leads to faster learning and higher final success on complex multitask benchmarks, and enables better zero-shot transfer of the learned policy to real robotic tasks including unseen tasks. 

\section*{Acknowledgement}
The authors would like to thank Lingfeng Sun, Xiaoqiang Yan, Jinghuan Shang, Laura Herlant, George Konidaris for helpful discussions and feedbacks.
\bibliographystyle{IEEEtran}
\bibliography{example}

\section{Limitations}
While our proposed data-efficient multitask DAgger framework demonstrates significant improvements in sample efficiency and sim-to-real transfer, it possesses several limitations.

Firstly, the method's effectiveness hinges on the availability and quality of pre-trained, task-specific expert policies. Although these experts are trained on state representations for efficiency, the initial cost associated with training $N$ distinct experts via reinforcement learning can still be considerable, particularly as the number of tasks $N$ grows very large. Our approach optimizes the distillation process but does not eliminate this prerequisite expert training phase.

Secondly, although our method achieved better zero-shot sim-to-real transfer compared to baselines, a performance gap between simulation and reality persists, especially for the out-of-domain task. This suggests that challenges related to domain shift are mitigated but not fully resolved by the adaptively sampled data distribution alone, and further techniques like domain randomization or incorporating minimal real-world data might be necessary for higher-fidelity transfer.

Finally, the current implementation has been validated on benchmarks with up to 36 tasks. While the scheduling mechanism is designed to be scalable, the practical overhead of managing a very large number of experts and the computational cost of the scheduling step itself might become factors when scaling to hundreds or thousands of diverse tasks.

\section{Algorithm Details}
\label{app:alg}
The process iterates through the following steps (see Algorithm~\ref{alg:multitask-dagger}):

\noindent \textbf{Initialization:} Train an initial policy $\pi_G^{(0)}$ using standard behavioral cloning (BC) on a small initial dataset $D^{(0)} = \bigcup_{i=1}^N D_i^{(0)}$ containing a few expert demonstrations for each task:
\begin{equation}
\small
\label{eq:bc-loss-generalist}
    \pi_G^{(0)} = \arg\min_{\pi_G} \sum_{i=1}^N \sum_{(o, a^*, z_i) \in D_i^{(0)}} \mathcal{L}_{\text{BC}}\!\bigl(\pi_G(a \mid o, z_i),\, a^*\bigr),
\end{equation}
where $\mathcal{L}_{\text{BC}}$ is the imitation loss (e.g., Mean Squared Error for continuous actions). Set the initial DAgger mixing coefficient $\epsilon_0$ (probability of using the learner's action during data collection). Initialize Kalman filter states (belief and uncertainty about success probability) $\{\hat{p}^i_0, P^i_0\}_{i=1}^N$ for each task (e.g., $\hat{p}^i_0=0.5, P^i_0=0.25$).

\noindent \textbf{Iteration $k=1, \dots, K$:}
\begin{enumerate}
    \item \textbf{Train Policy:} Update the generalist policy $\pi_G^{(k)}$ by training on the current aggregated dataset $D^{(k-1)}$ using Eq.~\eqref{eq:bc-loss-generalist}. During training, record the initial ($L^i_{k-1, \text{start}}$) and final ($L^i_{k-1, \text{end}}$) imitation loss specific to each task $i$ on a validation set or the training data itself. Calculate the performance gain $g^i_{k-1} = \max(0, L^i_{k-1, \text{start}} - L^i_{k-1, \text{end}})$.

    \item \textbf{Schedule Data Allocation:} Based on metrics from the \emph{previous} iteration's data collection (iteration $k-1$) and the training process just completed, determine the number of new expert trajectories $n_k^i$ to collect for each task $i$ in the current iteration $k$. This crucial step uses the performance-aware scheduler detailed in Section~\ref{subsec:scheduler}, ensuring $\sum_i n_k^i \approx B$, where $B$ is the total demonstration budget per iteration.

    \item \textbf{DAgger Data Collection:} For each task $i$, execute $\pi_G^{(k)}$ in the simulator for $n_k^i$ episodes. During these rollouts, with probability $\epsilon_{k-1}$, use the action $a_t = \pi_G^{(k)}(o_t, z_i)$; otherwise use the expert action $a_t^* = \pi_i^*(s_t)$. Critically, \emph{always record the expert label} $a_t^*$ paired with the observation $o_t$. Collect the resulting demonstration tuples $(o_t, a_t^*, z_i)$ into a temporary set $\Delta D_k^i$. Also, measure the raw success rate $\bar{p}_k^i$ of $\pi_G^{(k)}$ during these $n_k^i$ rollouts (or a fixed number of evaluation rollouts if $n_k^i=0$). This $\bar{p}_k^i$ will be used for scheduling in the \emph{next} iteration ($k+1$).

    \item \textbf{Update Dataset and Parameters:} Aggregate the newly collected data: $D^{(k)} = D^{(k-1)} \cup \bigcup_{i=1}^N \Delta D_k^i$. Decay the DAgger mixing coefficient: $\epsilon_k = \max(\epsilon_{k-1} - \Delta \epsilon, \epsilon_{\min})$.
\end{enumerate}

This loop continues until a maximum number of iterations $K$ is reached or a performance criterion is met. The key innovation lies in Step 2, the scheduler, which we detail next.

\section{Two important techniques of Performance-aware Scheduling}
\paragraph{Adaptive Kalman Filtering of Success Probabilities.}
Raw success probabilities $\bar{p}^i_{k-1}$ measured during DAgger collection can be unreliable for scheduling. The stochasticity of the policy and environment introduces noise, and estimates from a small number of rollouts $n_{k-1}^i$ suffer from high variance. To obtain a more stable estimate of the true success probability, $\hat{p}^i$, we use a Kalman filter for each task $i$.

The filter recursively updates its estimate $\hat{p}^i$ and associated uncertainty $P^i$. The process involves two steps:
\begin{enumerate}
    \item \textbf{Predict:} We assume the true success rate is stable between iterations, so the predicted probability is the last estimate, $\hat{p}^i_{k-1|k-2} = \hat{p}^i_{k-2|k-2}$. Our uncertainty in this prediction grows by a small process noise $Q$: $P^i_{k-1|k-2} = P^i_{k-2|k-2} + Q$.

    \item \textbf{Update:} The prediction is corrected using the new measurement $\bar{p}^i_{k-1}$. The degree of correction is determined by the Kalman gain $K^i_{k-1}$, which balances the predicted uncertainty against the measurement noise $R^i_{k-1}$. We make the measurement noise adaptive: $R^i_{k-1} = R_0 / (n_{k-1}^i + 1)$, reflecting higher confidence in estimates derived from more data. A lower measurement noise (more data) leads to a higher Kalman gain, placing more trust in the new measurement.
\end{enumerate}
The standard update equations are then applied:
\begin{align}
    K^i_{k-1} &= \frac{P^i_{k-1|k-2}}{P^i_{k-1|k-2} + R^i_{k-1}} \\
    \hat{p}^i_{k-1|k-1} &= \hat{p}^i_{k-1|k-2} + K^i_{k-1}\big(\bar{p}^i_{k-1} - \hat{p}^i_{k-1|k-2}\big) \\
    P^i_{k-1|k-1} &= (1 - K^i_{k-1}) \, P^i_{k-1|k-2}
\end{align}
The filter's output, $\hat{p}^i_{k-1} = \hat{p}^i_{k-1|k-1}$, provides a smoothed success probability for scheduling, mitigating the effects of noise and high-variance estimates. While the standard Kalman filter assumes Gaussian noise, and success rates are binomial, it serves as a simple and effective approximation for our purposes.
\paragraph{Rank Normalization of Performance Metrics.}
Empirically, we found that directly using the two metrics without normalization can lead to some overly tilted data distribution. For example, if one hard task got 0 success rate and the others got over 50\% success rate, almost all the data are collected from that hard task for the next iteration. To ensure robustness regardless of the chosen scheduling metric ('need' derived from $1-\hat{p}^i_{k-1}$ or 'gain' $g^i_{k-1}$), which may have different scales and distributions, we employ rank normalization. For the selected metric $m$ (either need or gain), let $\{m^1, \dots, m^N\}$ be the values across the $N$ tasks. Let $\text{rank}(m^i)$ denote the rank of task $i$ based on this metric (e.g., rank 1 for lowest value, rank $N$ for highest value). The rank-normalized value $\tilde{m}^i$ is computed as:
\begin{equation}
\label{eq:rank_norm}
    \tilde{m}^i = \frac{\text{rank}(m^i) - 1}{N - 1}, \quad \text{for } N > 1 \quad (\text{and } \tilde{m}^i = 0 \text{ if } N=1).
\end{equation}
This scales the ranks to the range $[0, 1]$, preserving the relative ordering while making the metrics scale-invariant and robust to outliers. We compute the normalized need $\tilde{d}^i_{k-1}$ (ranking $1-\hat{p}^i_{k-1}$) and normalized gain $\tilde{g}^i_{k-1}$ (ranking $g^i_{k-1}$) using this method.

We then take the outputs from normalization and use them as a priority score for each task $i$. We convert these scores into allocation probabilities using softmax and decide the number of trajectories allocated for each task by taking the product of this probability and the budget for the total number of trajectories to be allocated at each round. This adaptive mechanism dynamically focuses expert queries on tasks that are most likely to benefit, driving data efficiency.

\section{Implementation details}
Our method, which we refer to as MTDAgger (-PG or -TN), extends the standard DAgger framework by incorporating a performance-aware data collection schedule. As detailed in Section~\ref{sec:method}, we use a Kalman filter to estimate the success rate ($p_i$) for each task $i$, or use performance gain from training, $\Delta L_i$, that informs a scoring function. We use rank normalization on these metrics to calculate task scores. These scores determine the allocation of data collection budget for the next DAgger round using a temperature-controlled softmax function, prioritizing tasks estimated to benefit most from additional expert demonstrations. Key parameters for the scheduling mechanism (e.g., KF process/measurement variance, scoring weights $\alpha, \gamma$, softmax temperature) were set based on preliminary experiments and are detailed in the Appendix. The DAgger mixing parameter $\epsilon$ was annealed from 0.5 towards 0 over the rounds to transition from expert to learner actions.

We convert the normalized score into allocation probabilities using a temperature-controlled softmax:
\begin{equation}
\label{eq:softmax_alloc}
    \rho^i_k = \frac{\exp(\text{score}^i_{k-1} / T)}{\sum_{j=1}^N \exp(\text{score}^j_{k-1} / T)},
\end{equation}
where $T > 0$ is the temperature hyperparameter controlling the sharpness of the allocation, and $\rho^i_k$ represents the probability mass allocated to task $i$ in iteration $k$. The number of demonstrations allocated to task $i$ for the current iteration $k$ is then:
\[
    n_k^i = \max\left( n_{\min},\rho^i_k \cdot B \right),
\]
with minor adjustments to ensure the total allocation $\sum_i n_k^i$ approximately equals the budget $B$. We enforce a minimum allocation $n_{\min} \ge 1$ for tasks with non-zero scores. This adaptive mechanism dynamically focuses expert queries on tasks that are most likely to benefit, driving data efficiency.
\subsection{Multitask Policy Architecture for IsaacLab Drawer-opening} 
\label{subsec:architecture}
To handle diverse tasks within a single network, we employ a policy architecture that conditions on task identity. Each task $i$ is represented by a learned task embedding $z_i \in \mathbb{R}^d$, obtained by passing a one-hot task index through a trainable embedding layer. The policy $\pi_G(a \mid o, z_i)$ takes the current observation $o$ (e.g., robot proprioception and point cloud features) and the task embedding $z_i$ as input.

Our network consists of:
\begin{enumerate}
    \item An observation encoder $f_{\text{enc}}(o)$ (e.g., a PointNet-based module for point clouds, potentially combined with an MLP for proprioceptive state) that extracts a latent representation $h_o$.
    \item A fusion module that combines the observation representation $h_o$ with the task embedding $z_i$, for example, by concatenation: $h_{\text{fused}} = [h_o, z_i]$.
    \item A policy head $f_{\text{policy}}(h_{\text{fused}})$ (typically an MLP) that outputs the action $a$.
\end{enumerate}
The shared encoder $f_{\text{enc}}$ learns features relevant across all tasks, while the task embedding $z_i$ allows the policy head $f_{\text{policy}}$ to specialize its behavior for the specific requirements of task $i$. This architecture balances generalization and task-specific adaptation effectively for the visuomotor control tasks considered.

\begin{table*}[h]
\small
    \centering
    \begin{tabular}{cccccccccc}
        \hline
        Hyperparameters & Batch size & epsilon & epsilon decay &min epsilon & Q & $R_0$ & learning rate & T & $n_{min}$ \\ \hline
        MetaWorld & 1024 & 0.5 & 0.5 & 0 & 0.03 & 0.5 & 3e-4 & 0.5 & 1 \\
        IsaacLab & 256 & 0.5 & 0.5 & 0 & 0.03 & 0.5 & 1e-4 & 0.5 & 5 \\ \hline
    \end{tabular}
    \caption{Hyperparameters we used in our experiments. Although the benchmarks are completely different, the algorithm-related hyperparameters are basically the same.}
    \label{tab:hyperpara}
\end{table*}
\newpage
\section{Visualization of the simulated drawers}

\begin{figure}[!htb]
    \centering
    \includegraphics[width=0.99\linewidth]{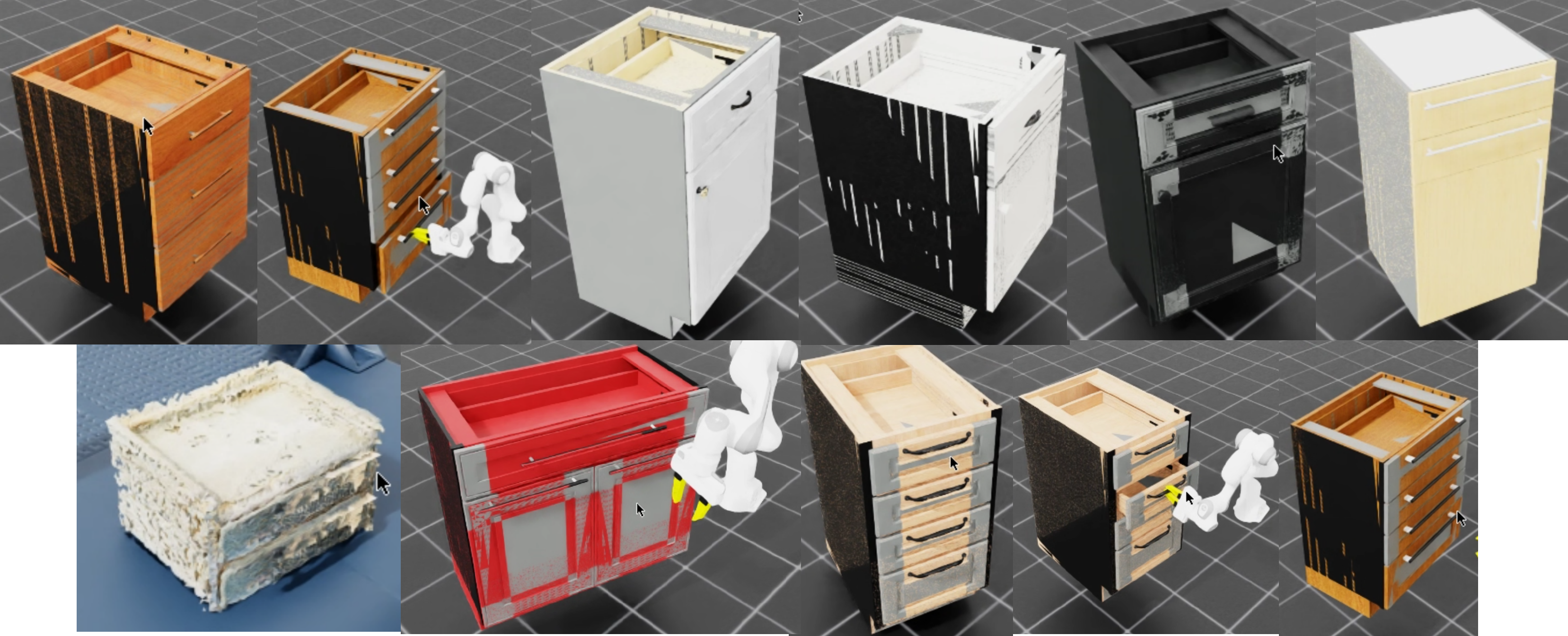}
    \caption{Visualization of the 11 drawers we trained on in the experiments.}
    \label{fig:drawers}
\end{figure}

\begin{figure}[!htb]
    \centering
    \includegraphics[width=0.3\linewidth]{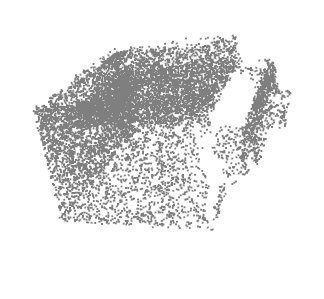}
    \includegraphics[width=0.3\linewidth]{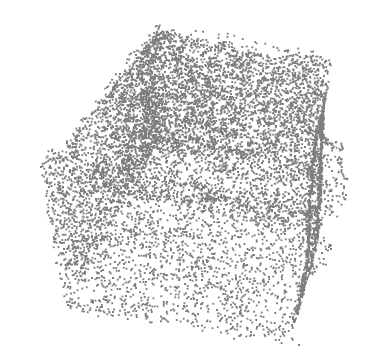}
    \includegraphics[width=0.3\linewidth]{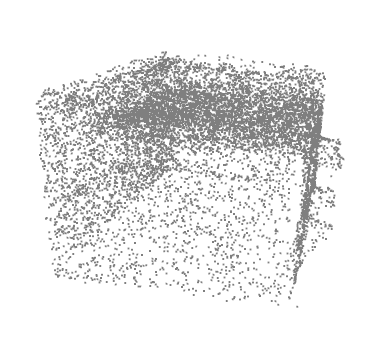}
    \caption{Examples of the point-cloud observations for real robot experiments.}
    \label{fig:pcd}
\end{figure}

\section{Other Metrics we have tried in the experiments}
Besides the two metrics -- Task Need and Performance Gain, that we have found to be useful in the experiments, in practice we have tried a few other metrics that didn't end up working well. These include:
\begin{itemize}
    \item KL divergence between the output distribution of the multitask policy and the expert policies
    \item Normalized final/initial loss of each task during each training round
    \item Variance of the normalized loss of each task during each training round
\end{itemize}

\end{document}